\documentclass[]{template}

\usepackage[utf8]{inputenc}             
\usepackage[T1]{fontenc}                
\usepackage{url}                        
\usepackage{booktabs}                   
\usepackage{multirow}
\usepackage{colortbl}
\usepackage{multicol}
\usepackage{amsfonts}                   
\usepackage{nicefrac}                   
\usepackage{microtype}                  
\usepackage[dvipsnames]{xcolor}         

\usepackage{latexsym}

\usepackage{graphicx}
\usepackage{float}
\usepackage{subcaption}
\usepackage{wrapfig}
\usepackage{lipsum}

\usepackage{bm}

\usepackage{tabularx} 
\usepackage{ragged2e} 
\newcolumntype{L}{>{\RaggedRight\hangafter=1\hangindent=0em}X}

\usepackage{enumitem}

\usepackage{amsmath}
\usepackage{amssymb}
\usepackage{mathtools}
\usepackage{amsthm}

\usepackage{hyperref}
\usepackage{url}
\usepackage{booktabs}
\usepackage{multirow}
\usepackage{graphicx}
\usepackage[table]{xcolor}
\usepackage{amssymb}
\usepackage{wasysym}
\usepackage{wrapfig}
\usepackage{float}
\usepackage{subcaption}

\setboolean{logo}{true}    

\usepackage[linesnumbered,ruled,vlined]{algorithm2e}

\hypersetup{
    colorlinks=true,
    linkcolor=red,
    citecolor=Cerulean,
    filecolor=magenta,      
    urlcolor=magenta,
}

\usepackage[capitalize,noabbrev]{cleveref}
\crefname{section}{§}{§§}
\Crefname{section}{§}{§§}

\usepackage{calligra}
\DeclareMathAlphabet{\mathcalligra}{T1}{calligra}{m}{n}

\usepackage{pifont}

\theoremstyle{plain}

\theoremstyle{definition}

\theoremstyle{remark}

\renewcommand{\paragraph}[1]{\vspace{1mm}\noindent\textbf{#1}}

\DeclareCaptionLabelFormat{cont}{#1~#2\alph{ContinuedFloat}}
\usepackage[most]{tcolorbox}
\tcbset{
  promptbox/.style={
    top=10pt,
    colback=lightgray!20,
    colframe=Black,
    colbacktitle=NavyBlue,
    enhanced,
    center,
    attach boxed title to top center={yshift=-0.1in,xshift=0.0in},
    boxed title style={boxrule=0pt,colframe=white,},
  }
}
\newtcolorbox{promptbox}[2][]{promptbox, title=#2,#1}
\tcbset{
  takeawaybox/.style={
    top=10pt,
    colback=lightgray!20,
    colframe=Black,
    colbacktitle=BurntOrange,
    enhanced,
    center,
    attach boxed title to top center={yshift=-0.1in,xshift=0.0in},
    boxed title style={boxrule=0pt,colframe=white,},
  }
}
\newtcolorbox{takeawaybox}[2][]{takeawaybox, title=#2,#1}
\tcbset{
  observationbox/.style={
    top=10pt,
    colback=lightgray!20,
    colframe=Black,
    colbacktitle=YellowGreen,
    enhanced,
    center,
    attach boxed title to top center={yshift=-0.1in,xshift=0.0in},
    boxed title style={boxrule=0pt,colframe=white,},
  }
}
\newtcolorbox{observationbox}[2][]{observationbox, title=#2,#1}

\usepackage{xspace}

\newcommand\blfootnote[1]{%
  \begingroup
  \renewcommand\thefootnote{}\footnote{#1}%
  \addtocounter{footnote}{-1}%
  \endgroup
}

\definecolor{gainBlue}{RGB}{0,80,200}
\newcommand{\gain}[1]{\,\textcolor{gainBlue}{\scriptsize$\uparrow$#1}}
\newcommand{\scoregain}[2]{%
    \shortstack[c]{#1\\[-1pt]\gain{#2}}%
}
\newcommand{\scoreonly}[1]{%
    \shortstack[c]{#1\\[-1pt]\vphantom{\gain{0.0000}}}%
}

\usepackage{CJK}

\title{InstantFusion: A Shared Latent Interface for Heterogeneous Image Generators}

\author[1,2]{Xuanlang Dai$^{*}$}
\author[2,3]{Jiazi Bu$^{*}$}
\author[3]{Yujie Zhou}
\author[2,4]{Yuhong Liu}
\author[2,4]{Beichen Zhang}
\author[2,3]{Penghui Yang}
\author[2,3]{Ziyu Liu}
\author[2]{Yuhang Zang$^{\dagger}$}

\affil[1]{Fudan University}
\affil[2]{Shanghai AI Laboratory}
\affil[3]{Shanghai Jiao Tong University}
\affil[4]{The Chinese University of Hong Kong}

\begin{abstract}
Diffusion/flow models have emerged as the de facto paradigms for image generation, giving rise to a diverse family of modern generators with increasingly heterogeneous architectures and representations.
Despite their remarkable generation capabilities, the cross-model alignment of latent representations and the consistency of their underlying denoising dynamics remain underexplored.
In this work, we uncover an intriguing cross-model compatibility:
independently trained generators exhibit similar latent spatial organization, while directly handing off intermediate states between unaligned models can still preserve coarse visual content.
Motivated by these observations, we propose \textbf{InstantFusion}, a lightweight shared latent interface that connects the denoising trajectories of heterogeneous image generators.
Specifically, to enable flexible cross-model state translation, model-specific Latent AutoEncoders (\textbf{LAE}s) are introduced to project intermediate states from different generators into a shared latent space and decode them into the latent space of a target generator. This shared space is constrained by two complementary objectives: aligning intermediate states derived from the same image at matched noise levels and synchronizing their corresponding local denoising velocities.
Once trained, InstantFusion allows denoising to seamlessly transition across frozen generators without requiring a dedicated mapping for every model pair.
The versatility of InstantFusion is demonstrated across acceleration, multi-preference composition, cross-architecture on-policy distillation, and cross-model composition, achieving competitive performance across these diverse settings.

\end{abstract}

\begin{document}

\blfootnote{$^{*}$ Equal contribution.}
\blfootnote{$^{\dagger}$ Corresponding author: Yuhang Zang
(\href{mailto:zangyuhang@pjlab.org.cn}{zangyuhang@pjlab.org.cn}).}
\blfootnote{$^{\ddagger}$ Code is available at
\url{https://github.com/InternLM/InstantFusion}.}

\maketitle

\section{Introduction}

Diffusion and flow models have emerged as the dominant paradigms for modern image generation~\citep{song2023consistency,esser2024scaling,wu2025qwen,ho2020denoising,song2020score,lipman2022flow}, enabling increasingly powerful generators with diverse architectural designs. 
Meanwhile, the model landscape has grown substantially more heterogeneous: generative models differ fundamentally in latent representations and conditioning mechanisms~\citep{esser2024scaling,wu2025qwen}, network architectures and parameter scales~\citep{cai2025z,blackforestlabs2025flux2,podell2024sdxl}, and training or distillation recipes~\citep{salimans2022progressive,song2023consistency,luo2023latent,yin2024one}. Despite these structural differences, generative models ultimately learn to map noise onto closely related natural-image distributions under textual conditioning. However, they are typically studied and deployed as isolated systems. This isolation leaves it largely unexplored whether their latent representations admit meaningful cross-model alignment and whether their underlying generative dynamics remain compatible across distinct models.

\begin{figure}[t]
    \centering
    \includegraphics[width=0.95\linewidth]{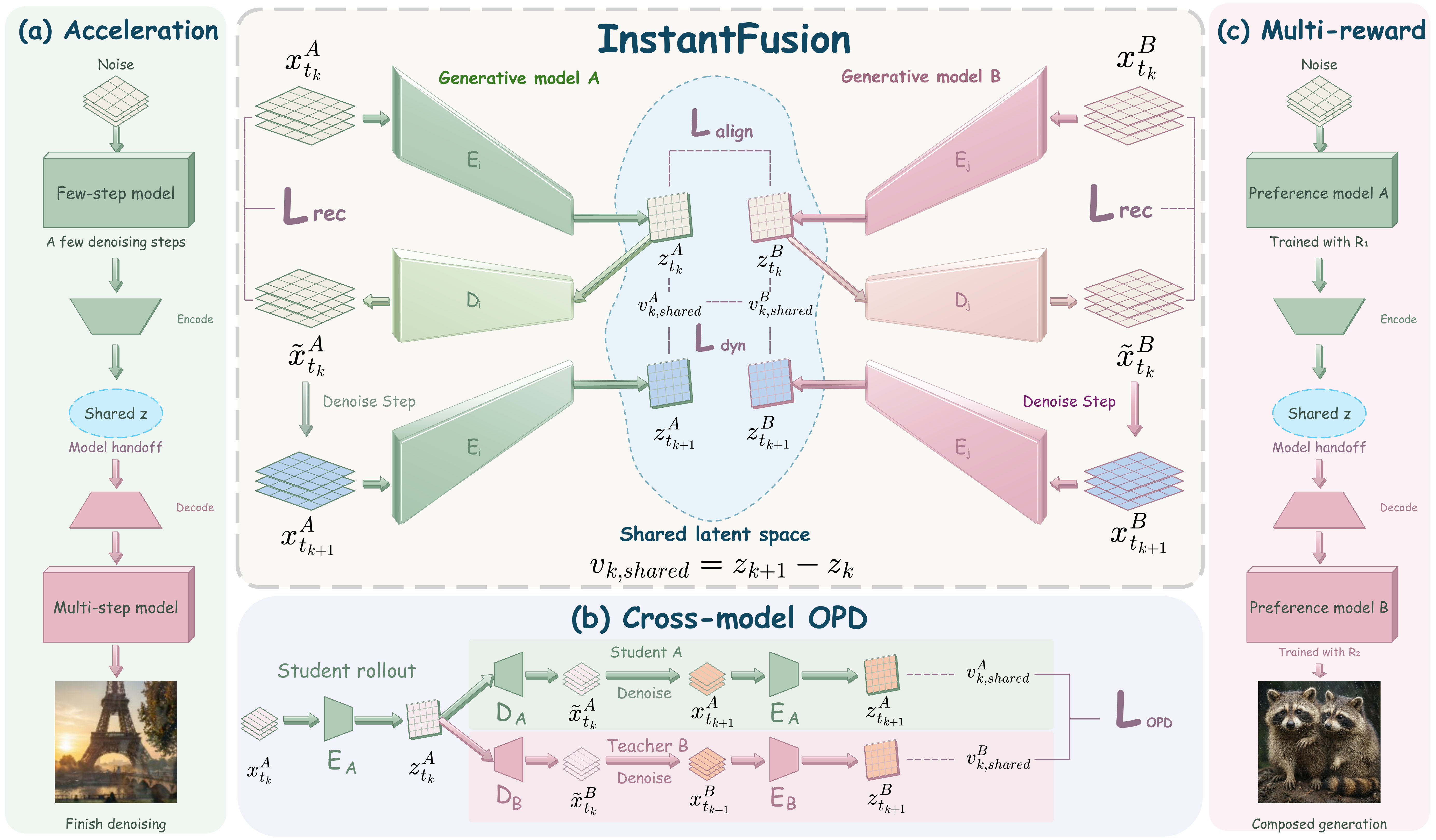}
    \caption{
    Overview of InstantFusion.
    During training, model-specific LAEs map the intermediate
    states of frozen generators into a shared representation
    space, where both latent states and denoising dynamics are
    aligned.
    During inference, a sender state is translated through the
    shared space into the receiver's latent space, and an adaptive
    criterion determines when the receiver takes over denoising.
    }
    \label{fig:method_overview}
\end{figure}

In this work, we observe two indications of compatibility among independently trained generators: 
First, despite differences in latent dimensionality, tokenization, and parameterization, their intermediate states often exhibit closely related spatial organization (Fig.~\ref{fig:observation}(a)). 
Second, directly handing an intermediate denoising state from one model to another, 
without any learned alignment, 
typically introduces artifacts but can preserve the main subject and coarse spatial layout of the generated image (Fig.~\ref{fig:observation}(b)). 
These observations suggest that heterogeneous generators may encode partially shared visual structure and that their denoising processes are not entirely disconnected. 
This naturally raises a more practical question: \textit{can such partial compatibility be turned into a reusable interface for transferring both latent states and denoising dynamics across heterogeneous latent spaces?}

Existing approaches provide only partial answers to this question. 
Cross-model latent alignment methods establish connections between individual states from different models~\citep{asperti2023comparing,maiorca2023latent,badrinath2024all}, 
but they primarily focus on model-pair-specific mappings, 
lacking both a shared representation space that can accommodate multiple generators and an explicit treatment of denoising dynamics. 
Meanwhile, trajectory stitching methods allow various denoisers to take over at different stages of sampling,
but generally assume that the participating models operate within the same latent space~\citep{pan2025t}. 
Therefore, transferring denoising trajectories among generators with heterogeneous latent spaces remains an open problem.

Motivated by the analysis above, we introduce \textbf{InstantFusion}, 
a lightweight shared latent interface that connects the denoising trajectories of heterogeneous image generators (Fig.~\ref{fig:method_overview}).
Each frozen generator uses a model-specific Latent AutoEncoder (\textbf{LAE})
that maps its intermediate states into a shared representation 
and reconstructs them in the model's native latent space. 
To learn this interface, we construct paired intermediate states from the same image at matched noise levels
and align their encoded representations. 
Since state correspondence alone is insufficient for reliable trajectory transfer,
additional alignment of the projected local denoising velocities is enforced 
so that translated states follow the receiver's denoising trajectory.
At inference, a sender state is encoded by the sender's encoder and decoded by the receiver's decoder into the receiver's latent space.
This design allows new generators to be incorporated through their own LAEs, 
enabling flexible cross-model transitions without dedicated pairwise mappings.

InstantFusion provides two fundamental interface capabilities: faithful reconstruction and cross-model state transfer. The latter directly enables cross-model composition: composing Qwen-Image~\citep{wu2025qwen} with SD3 or FLUX improves GenEval-v2~\citep{kamath2025geneval} by 13 points and CVTG-2K~\citep{du2025textcrafteraccuratelyrenderingmultiple} by 0.18--0.28 over the corresponding base models. Building on interface-level capabilities, InstantFusion supports three downstream applications shown in Fig.~\ref{fig:method_overview}: generation acceleration, cross-architecture on-policy distillation, and multi-preference composition. It achieves a $1.8$--$3.3\times$ generation speedup while maintaining strong perceptual quality, balances multiple specialized preferences with competitive GenEval-v2 and HPSv2~\citep{wu2023human} performance, and enables on-policy knowledge transfer from Qwen-Image to the architecturally distinct SD3. Experiments demonstrate that InstantFusion serves as a general interface for capability transfer and composition across heterogeneous image generators.

Our contributions are threefold:
(1) Empirical evidence reveals partial compatibility among independently trained image generators, reflected in similar latent spatial organization and the preservation of coarse visual content under direct, zero-shot unaligned handoffs.
(2) InstantFusion provides a lightweight shared latent interface that aligns intermediate representations and local denoising dynamics, enabling seamless denoising trajectory transfer across heterogeneous latent spaces without pair-specific mappings.
(3) We demonstrate InstantFusion across generation acceleration, multi-preference composition, cross-architecture on-policy distillation, showing its potential as a general interface for capability transfer among heterogeneous generators.

\section{Related Work}
We discuss key studies here and defer others to the appendix~\ref{app:relatedwork}.

\textbf{Text-to-Image Diffusion Models}. Diffusion and flow-based models have become the prevailing frameworks for text-to-image generation~\citep{ho2020denoising,lipman2022flow}. Modern systems commonly perform denoising in learned latent spaces, combining pretrained autoencoders with text-conditioned denoisers~\citep{rombach2022high,podell2024sdxl}. Within this general framework, model designs have become increasingly diverse. Recent generators, including SD3, FLUX.1, Qwen-Image, and Z-Image, differ substantially in their latent parameterizations, denoising architectures, and training objectives~\citep{esser2024scaling,blackforestlabs2025flux2,wu2025qwen,cai2025z}. Yet these independently developed generators are typically studied in isolation, leaving the relationships among their intermediate representations and denoising dynamics largely unexplored.

\textbf{Cross-Model Latent Alignment}. A related line of work investigates whether independently trained generative models develop structurally compatible latent representations. These studies establish cross-model correspondences through linear mappings, relative representations, or closed-form transformations~\citep{asperti2023comparing,moschella2022relative,maiorca2023latent,badrinath2024all}. More recently, learned bridges have been proposed to connect incompatible teacher and student VAE latent spaces, enabling cross-space diffusion distillation~\citep{nguyen2026cross}. Nevertheless, these approaches operate primarily on isolated latent states and do not model how corresponding states evolve along denoising trajectories. It therefore remains unclear whether static latent correspondence also implies compatibility between the underlying generative dynamics.

\section{Methods}

InstantFusion enables handoff between frozen image generators by aligning their latent representations and denoising dynamics. In Sec.~\ref{met:observation}, we identify spatial self-similarity and preserved coarse semantics across unaligned models. Building on this, Sec.~\ref{met:static} equips each generator with a model-specific LAE to align and reconstruct intermediate states, while Sec.~\ref{met:dynamic} synchronizes dynamic velocities to harmonize trajectories. For inference, Sec.~\ref{met:adaptive} introduces an adaptive strategy tracking predicted clean latents for optimal switching. Fig.~\ref{fig:method_overview} provides an overview of our full pipeline.

\begin{figure}[t]
    \centering
    \includegraphics[width=1\linewidth]{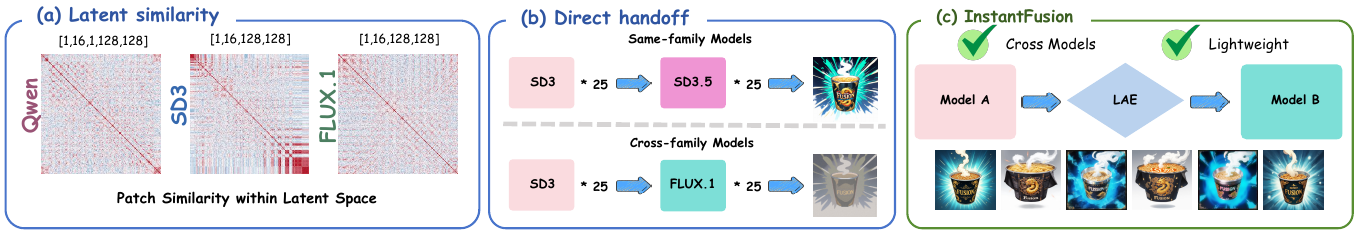}
    \caption{\textbf{Cross-model compatibility.}
    (a) Intermediate states from Qwen-Image, SD3, and FLUX.1 exhibit similar spatial self-similarity patterns.
    (b) Direct handoff between related models preserves generation quality, while cross-family handoff retains the main subject and coarse layout.
    (c) These observations motivate a lightweight interface for connecting heterogeneous generators.}
    \label{fig:observation}
\end{figure}

\subsection{Preliminaries}
\label{met:preliminaries}

\noindent\textbf{Flow Matching.}
Flow Matching (FM)~\citep{lipman2022flow} learns a vector field
that transports noise to data. Let
$\mathbf{x}_0\sim\pi_0$ be a clean latent drawn from the data
latent distribution $\pi_0$, and let
$\boldsymbol{\epsilon}\sim\mathcal{N}(\mathbf{0},\mathbf{I})$
be standard Gaussian noise. At time $t\in[0,1]$, the linear
interpolation is
\begin{equation}
    \mathbf{x}_t=(1-t)\mathbf{x}_0+t\boldsymbol{\epsilon},
\end{equation}
where $t=0$ denotes clean data and $t=1$ denotes noise.
The vector field $v_\theta(\mathbf{x}_t,t,c)$, parameterized
by $\theta$ and conditioned on prompt representation $c$,
is trained to predict the path velocity
$\boldsymbol{\epsilon}-\mathbf{x}_0$:
\begin{equation}
    \mathcal{L}_{\mathrm{FM}}
    =
    \mathbb{E}_{\mathbf{x}_0,c,\boldsymbol{\epsilon},t}
    \left[
        \left\|
            v_\theta(\mathbf{x}_t,t,c)
            -(\boldsymbol{\epsilon}-\mathbf{x}_0)
        \right\|_2^2
    \right].
\end{equation}
Here, $\|\cdot\|_2$ denotes the Euclidean norm.
Generation integrates the learned field from $t=1$ to $t=0$.

\subsection{Observation: Shared Structure Across Models}
\label{met:observation}

We identify two concrete signals of latent compatibility across independently trained generators. First, intermediate representations maintain equivalent spatial dimensions across model architectures, such as Qwen-Image ($[1,16,1,128,128]$) alongside SD3 and FLUX.1 ($[1,16,128,128]$), yielding similar spatial patterns that reflect comparable information density (Fig.~\ref{fig:observation}(a)). Second, halting SD3 partway through denoising and handing off the intermediate state to SD3.5 or FLUX.1 to complete the remaining steps produces contrasting outcomes: while same-family handoff yields a seamless image, cross-family handoff introduces local artifacts yet clearly preserves the central cup and its coarse layout (Fig.~\ref{fig:observation}(b)). These artifacts confirm that native latent spaces cannot be directly interchanged, whereas the preserved cup layout suggests that some coarse semantic structure is shared across the latent spaces. These insights directly motivate an interface that aligns and reconstructs intermediate states before harmonizing their denoising dynamics.

\subsection{Latent Alignment and Reconstruction}
\label{met:static}

\begin{figure}[t]
    \centering
    \includegraphics[width=1\linewidth]{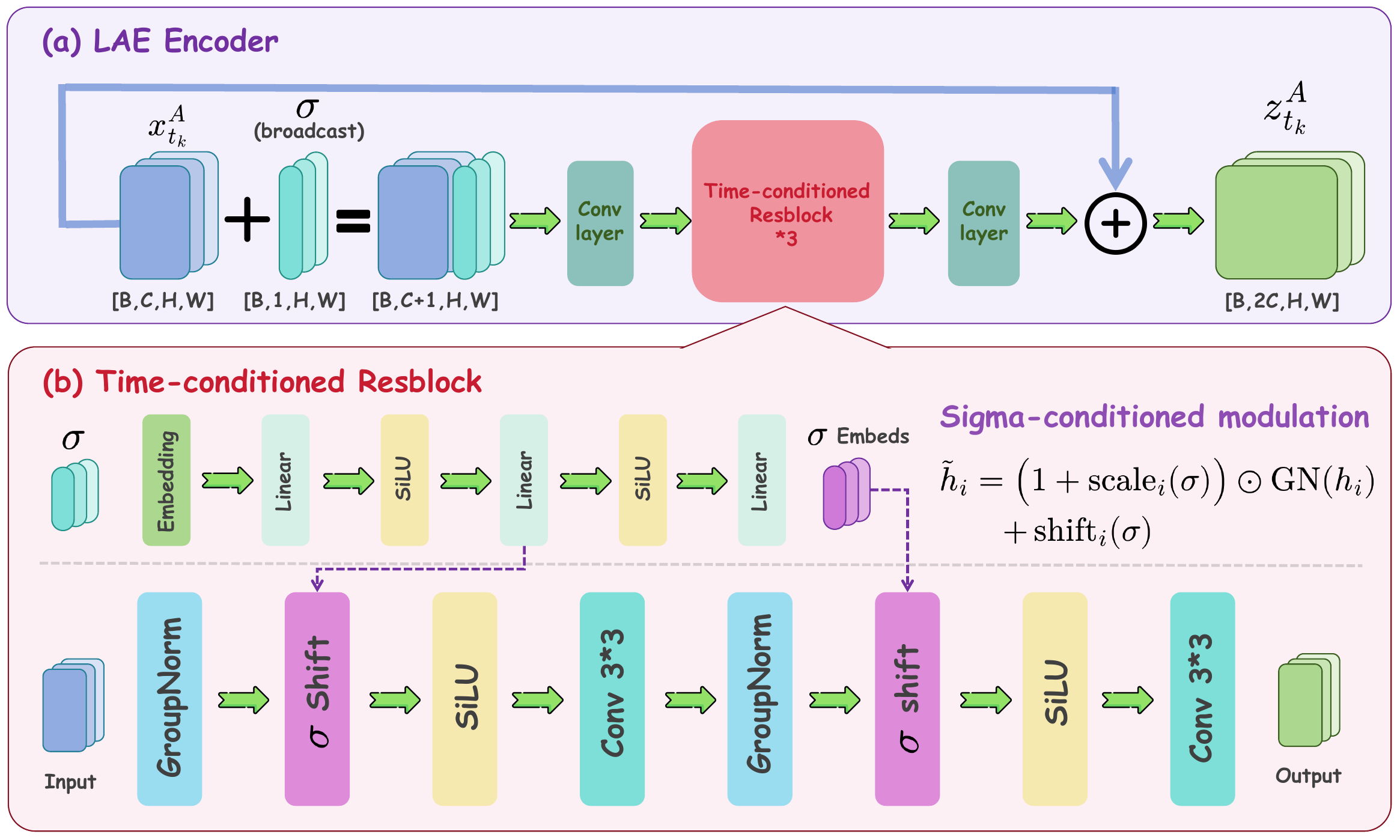}
    \caption{\textbf{Architecture of the LAE.}
(a) $\sigma$ denotes the scalar noise level associated with
denoising step $t_k$. It is spatially broadcast and concatenated with
the intermediate state $\mathbf{x}_{t_k}^{A}$. Convolutional layers and time-conditioned residual blocks then map the input to its encoded representation $\mathbf{z}_{t_k}^{A}$, with a skip connection preserving the original state.
(b) Details of the time-conditioned resblock: Each residual block embeds $\sigma$ and predicts scale and shift parameters that modulate group-normalized features, conditioning the transformation on the current denoising stage.}
    \label{fig:lae_architecture}
\end{figure}

Let $m\in\{1,\ldots,M\}$ index the $M\geq2$ frozen generators.
All generators use the same frozen variational autoencoder
(VAE). We encode an image into a clean latent $\mathbf{x}_0$
and construct a common noisy input $\mathbf{x}_t$ using the
same noise sample $\boldsymbol{\epsilon}$ and time $t$.

Each generator is equipped with a lightweight, time-conditioned
LAE consisting of an encoder $E_m$ and a decoder $D_m$, whose
architecture is illustrated in Fig.~\ref{fig:lae_architecture}.
\begin{equation}
    \mathbf{z}_t^m=E_m(\mathbf{x}_t,t),
    \qquad
    \widetilde{\mathbf{x}}_t^m=D_m(\mathbf{z}_t^m,t),
\end{equation}
Cross-model handoff requires the LAE encoders to map corresponding
intermediate states to a common representation that can be consumed
by any model's decoder. We enforce this correspondence with the
pairwise alignment loss, where $A$ and $B$ index distinct generators.
\begin{equation}
\mathcal{L}_{\mathrm{align}}
=
\frac{2}{M(M-1)}
\sum_{1 \leq A < B \leq M}
\left\lVert
\mathbf{z}_t^A-\mathbf{z}_t^B
\right\rVert_2^2 .
\end{equation}

\begin{figure}[t]
    \centering
    \includegraphics[width=1\linewidth]{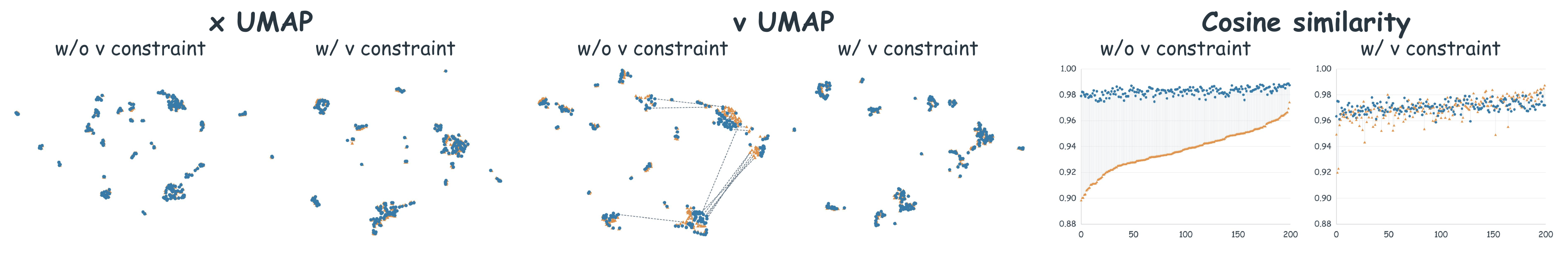}
    \caption{\textbf{Effect of denoising-dynamics alignment.}
Training without $\mathcal{L}_{\mathrm{dyn}}$ establishes correspondence between intermediate states $\mathbf{x}$, but leaves their associated denoising velocities $\mathbf{v}$ poorly aligned. Adding $\mathcal{L}_{\mathrm{dyn}}$ improves velocity alignment. Each line in the UMAP visualization connects the $\mathbf{v}$ feature points produced by different models when conditioned on the same prompt.}
    \label{fig:state_dynamics_clustering}
\end{figure}

\begin{figure}[t]
    \centering
    \includegraphics[width=1\linewidth]{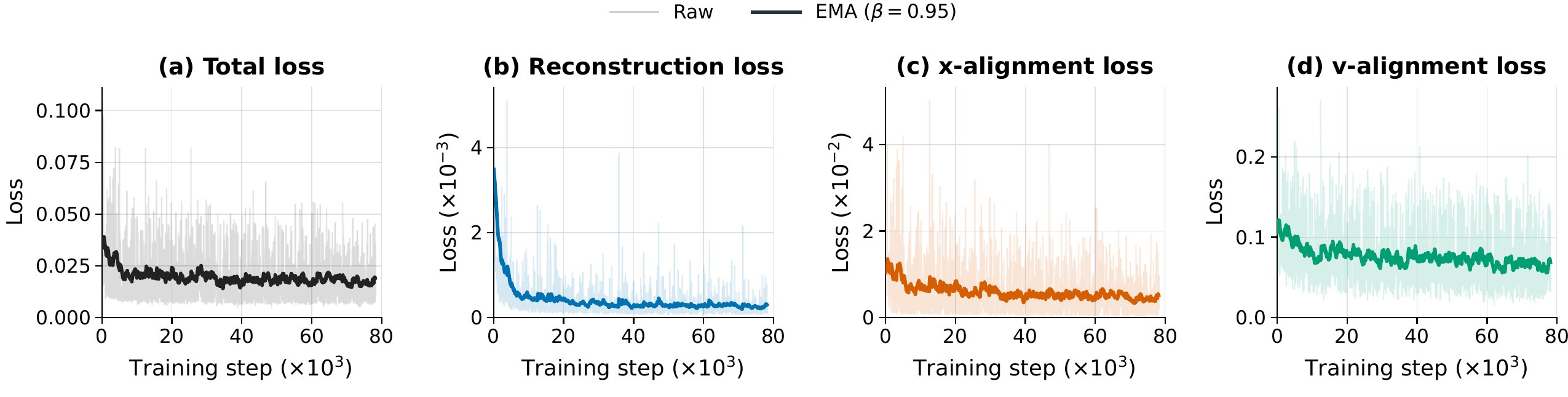}
    \caption{Training dynamics of the model-specific LAEs. The total loss and the three component losses ($\mathcal{L}_{\mathrm{rec}}$, $\mathcal{L}_{\mathrm{align}}$, and $\mathcal{L}_{\mathrm{dyn}}$) decrease steadily and converge during training.}
    \label{fig:training_losses}
\end{figure}

However, minimizing $\mathcal{L}_{\mathrm{align}}$ alone admits a trivial solution that satisfies the pairwise-distance objective while discarding all input-dependent information. We therefore introduce
a reconstruction loss to preserve the latent information required
for subsequent denoising
\begin{equation}
    \mathcal{L}_{\mathrm{rec}}
    =
    \frac{1}{M}
    \sum_{m=1}^{M}
    \|\widetilde{\mathbf{x}}_t^m-\mathbf{x}_t\|_2^2.
\end{equation}

where $\widetilde{\mathbf{x}}_t^m$ denotes the reconstructed latent obtained by passing $\mathbf{x}_t$ through model $m$'s LAE encoder and decoder. Moreover, to encourage comparable representation magnitudes across
models for the same image and timestep, we introduce a
norm-consistency regularization term
\begin{equation}
    \mathcal{L}_{\mathrm{reg}}
    =
    \frac{2}{M(M-1)}
    \sum_{1\leq A<B\leq M}
    \left(
        \|\mathbf{z}_t^A\|_2
        -
        \|\mathbf{z}_t^B\|_2
    \right)^2.
\end{equation}

\subsection{Denoising Dynamics Alignment}
\label{met:dynamic}


To examine whether latent state alignment also aligns denoising
dynamics, Fig.~\ref{fig:state_dynamics_clustering} analyzes shared states and their one-step
displacements using UMAP and cosine similarity.
The UMAP visualizations show that corresponding states from
different models cluster closely after state alignment,
whereas their denoising displacements remain separated.
Consistently, cross-model cosine similarity is high for
paired states but lower for their displacements.
This mismatch motivates explicitly aligning one-step
denoising displacements in the shared space.

Let $1=t_0>\cdots>t_K=0$ be a common sampling schedule with
$K$ denoising steps. For step $k\in\{0,\ldots,K-1\}$, define
$\Delta t_k=t_{k+1}-t_k$. At step $k$, generator $m$ applies its
frozen vector field $v_{\theta_m}$ to the reconstructed state,
conditioned on its own encoding $c^m$ of the shared prompt
\begin{equation}
    \mathbf{x}_{t_{k+1}}^m
    =
    \widetilde{\mathbf{x}}_{t_k}^m
    +\Delta t_k\,
    v_{\theta_m}
    (\widetilde{\mathbf{x}}_{t_k}^m,t_k,c^m),
\end{equation}
where $\mathbf{x}_{t_{k+1}}^m$ denotes the resulting
model-specific latent state. Its displacement in the shared
space is
\begin{equation}
    \Delta\mathbf{z}_k^m
    =
    E_m(\mathbf{x}_{t_{k+1}}^m,t_{k+1})
    -\mathbf{z}_{t_k}^m.
\end{equation}
Since all generators use the same time interval, we directly
align these displacements:
\begin{equation}
    \mathcal{L}_{\mathrm{dyn}}
    =
    \frac{2}{M(M-1)}
    \sum_{1\leq A<B\leq M}
    \operatorname{MSE}
    (\Delta\mathbf{z}_k^A,\Delta\mathbf{z}_k^B),
\end{equation}
Therefore, the LAEs are jointly optimized with
\begin{equation}
    \mathcal{L}
    =
    \mathcal{L}_{\mathrm{rec}}
    +\mathcal{L}_{\mathrm{align}}
    +\lambda_{\mathrm{reg}}\mathcal{L}_{\mathrm{reg}}
    +\lambda_{\mathrm{dyn}}\mathcal{L}_{\mathrm{dyn}},
\end{equation}

where $\lambda_{\mathrm{reg}}$ and $\lambda_{\mathrm{dyn}}$ denote
the relative weights assigned to norm regularization and dynamics
alignment. We use a small
$\lambda_{\mathrm{reg}}$ for training stability. The objective is averaged over sampled training states, while the generators remain
frozen and only the LAEs are optimized.

\subsection{Adaptive Model Handoff}
\label{met:adaptive}

The optimal handoff step varies across model pairs, making manual tuning impractical and unscalable as more generators are integrated. Crucially, executing a handoff too early in the trajectory proves ineffective because the sender model has yet to crystallize stable semantic structures from pure noise. To overcome this limitation, the handoff step is determined online, dynamically guided by the temporal consistency of the sender's predicted clean states across consecutive denoising steps.

For a transition from model $A$ to model $B$, let
$\mathbf{x}_{t_k}^A$ denote the current latent state of model $A$ at
inference step $k$. Its translation into the latent space of model
$B$ is

\begin{equation}
    \mathbf{x}_{t_k}^{A\rightarrow B}
    =
    D_B(E_A(\mathbf{x}_{t_k}^A,t_k),t_k),
\end{equation}

For any latent state $\mathbf{x}$ at time $t$ with conditioning
$c$, define model $m$'s clean-state prediction:

\begin{equation}
    \mathbf{x}_{0,k}
    = \mathbf{x}-t\,v_{\theta}(\mathbf{x},t_k,c).
\end{equation}

Here, $\mathbf{x}-t\,v_{\theta}(\mathbf{x},t_k,c)$ estimates
the clean latent under the linear flow path. We quantify the prediction stability by measuring the
agreement between its clean predictions at two consecutive steps:

\begin{equation}
S_k^{A}
=
\frac{
\left\|
\mathbf{x}_{0,k}^{A}
-
\mathbf{x}_{0,k-1}^{A}
\right\|_2
}{
\max\left(
\left\|\mathbf{x}_{0,k-1}^{A}\right\|_2,
\kappa
\right)
},
\qquad
\kappa=10^{-8}.
\label{eq:relative_clean_change}
\end{equation}

\begin{figure}[H]
    \centering
    \includegraphics[width=1\linewidth]{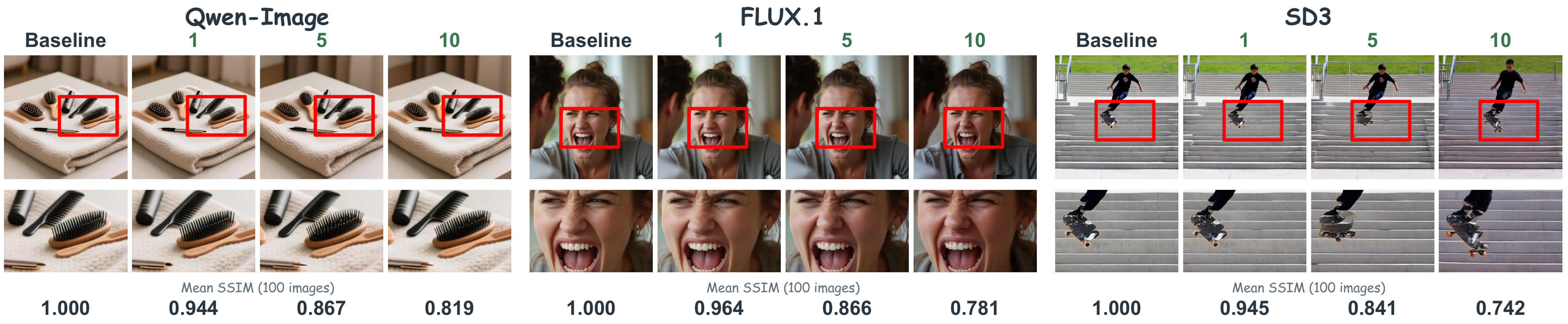}
    \caption{Qualitative comparison of Qwen-Image, FLUX.1, and SD3 outputs after 1, 5, and 10 repeated latent round trips. The original outputs before any round trip serve as baselines.}
    \label{fig:reconstruction_cases}
    \vspace{-1em}
\end{figure}

\begin{figure}[H]
    \centering
    \includegraphics[width=1\linewidth]{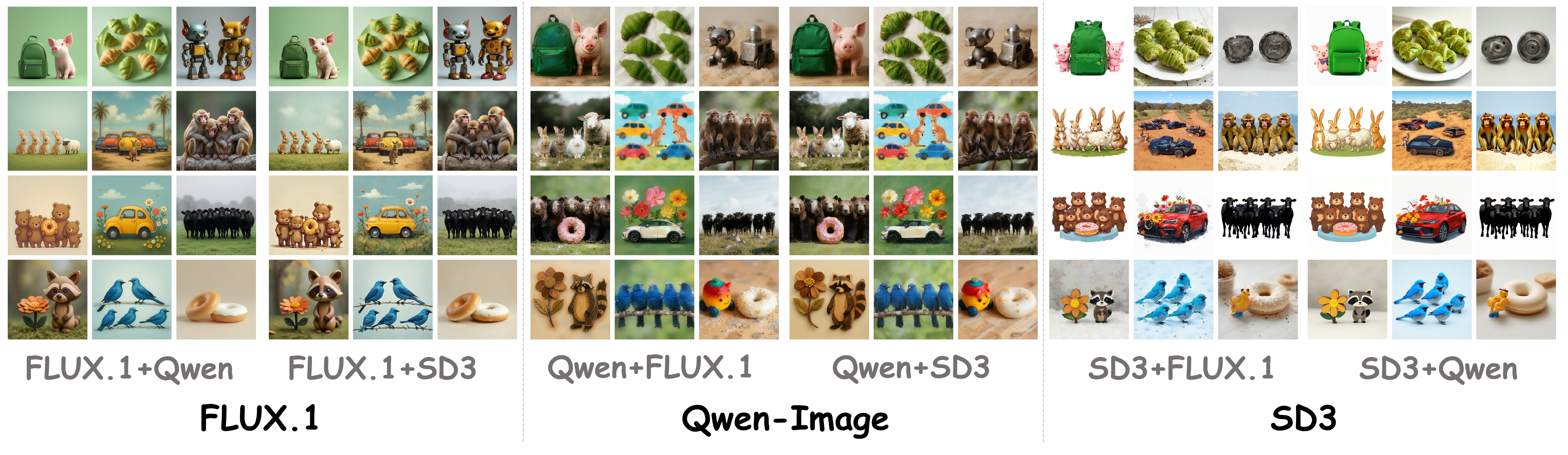}
    \caption{Qualitative results of cross-model latent transfer among heterogeneous generators.}
    \label{fig:cross_model_transfer}
    \vspace{-1em}
\end{figure}

\begin{figure}[H]
    \centering
    \includegraphics[width=1\linewidth]{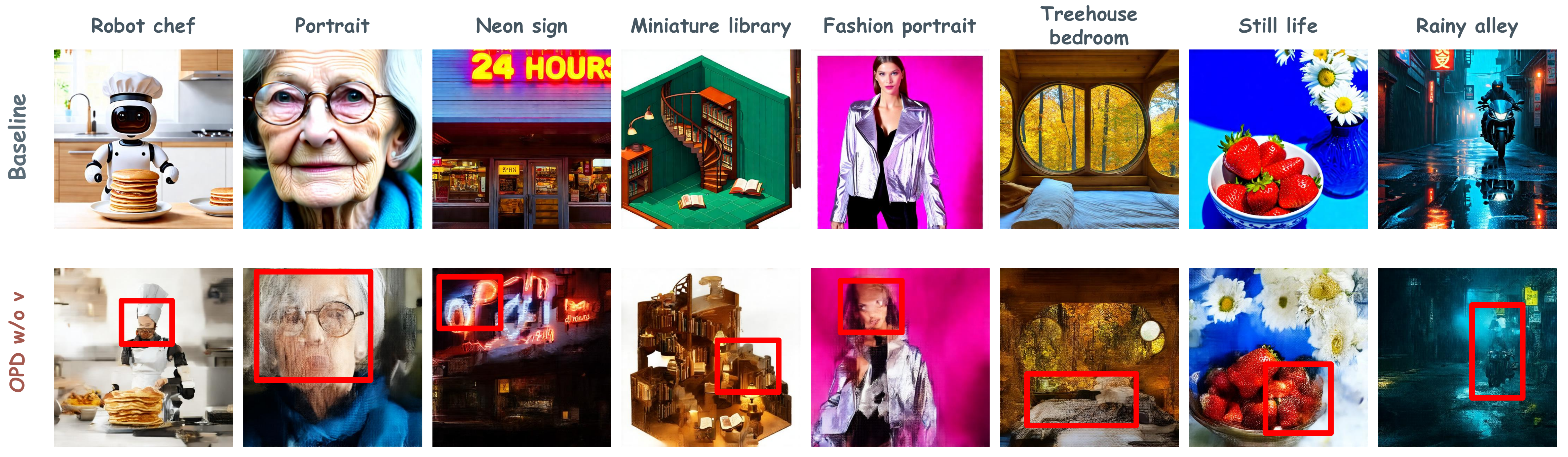}
    \vspace{-2em}
    \caption{Failure case of cross-model on-policy distillation after removing $\mathcal{L}_{\mathrm{dyn}}$.}
    \vspace{-1.5em}
    \label{fig:dynamics_alignment_failure}
\end{figure}

Given a similarity threshold $\gamma\in[-1,1]$, we select
the first qualifying step. The receiver then continues denoising from
$\mathbf{x}_{t_{k_{\mathrm{switch}}}}^{A\rightarrow B}$.

\begin{equation}
    k_{\mathrm{switch}}
    =
    \min\left\{
        k\in\{0,\ldots,K-1\}
        \;\middle|\;
        s_k^{A}\geq\gamma
    \right\}.
\end{equation}

\section{Experiments}


\subsection{Experimental Setup}
\label{sec:experimental_setup}

The experiments are conducted on four representative text-to-image generators: SD3~\citep{esser2024scaling}, Z-Image~\citep{cai2025z}, FLUX.1~\citep{blackforestlabs2025flux2}, and Qwen-Image (Qwen)~\citep{wu2025qwen}, covering diverse architectures and latent parameterizations. We use 50,000 image-text pairs from the VisPrompt dataset~\citep{yi2026flowinone} for training, while keeping all generator backbones frozen to optimize only the LAEs. Specifically, the LAEs are parameterized with 32 shared channels, 64 hidden channels, and a 128-dimensional $\sigma$ embedding, trained with an initial learning rate of $10^{-4}$. The reconstruction, alignment, dynamics-alignment, and norm regularization losses are balanced with weights of $1.0$, $1.0$, $0.1$, and $10^{-4}$, respectively. Further implementation details are provided in the appendix~\ref{app:details}. As demonstrated in Fig.~\ref{fig:training_losses}, all training objectives decrease steadily throughout optimization, confirming that the LAEs achieve accurate reconstruction while effectively aligning cross-model latent states and their corresponding denoising dynamics.

\subsection{Performance of LAE}
\label{sec:how}

\textbf{Reconstruction Fidelity}. 
We first jointly train one LAE per generator to project its native latent space into a shared representation. To verify how much information is preserved by the shared representation, clean latents are mapped into the shared space and reconstructed back into their native domains. As reported in Tab.~\ref{tab:reconstruction}, single-trip encoding achieves high-fidelity reconstruction (e.g., PSNR ${>}$ 31 dB, SSIM ${>}$ 0.94), demonstrating that the shared space retains sufficient detail, though performance degrades as trips increase. Qualitative results are illustrated in Fig.~\ref{fig:reconstruction_cases}. 

\textbf{Cross-Model Transfer}.
Beyond preserving native latents, the shared interface enables
cross-generator exchange of intermediate states. To assess this
capability, intermediate representations are translated via the shared
space to resume denoising across SD3, FLUX.1, and Qwen-Image. As detailed
in Tab.~\ref{tab:backbone_comparison}, guidance from Qwen-Image improves
the GenEval-v2 scores of SD3 and FLUX.1 by $12.92$ and $13.68$,
respectively, reaching absolute scores of $35.06$ and $36.54$. The
corresponding combinations also improve CVTG-2K by $0.2806$ and
$0.1803$, and the OneIG-Bench~\citep{chang2026oneig} average by
$0.0619$ and $0.0591$. Together with the qualitative results in
Fig.~\ref{fig:cross_model_transfer}, these gains demonstrate effective
state transfer across heterogeneous architectures.




\begin{table}[t]
\centering
\caption{
\textbf{Evaluation of the shared latent interface.}
(a) Reconstruction fidelity across repeated latent round-trips.
(b) Quantitative generation performance under cross-architecture on-policy distillation.
(c) Reconstruction fidelity degradation upon removing the reconstruction loss $\mathcal{L}_{\mathrm{rec}}$.
}
\label{tab:compact_evaluation}

\captionsetup[subtable]{
    font=small,
    labelfont=bf,
    justification=centering,
    skip=2pt
}

\begin{minipage}[t]{0.49\textwidth}
\centering

\subcaptionbox{
    Repeated latent round trips.
    \label{tab:reconstruction}
}[\linewidth]{
\resizebox{\linewidth}{!}{
\begin{tabular}{lcccc}
\toprule
\textbf{Shared Space}
& \textbf{Trips}
& \textbf{PSNR} $\uparrow$
& \textbf{SSIM} $\uparrow$
& \textbf{LPIPS} $\downarrow$ \\
\midrule

\multirow{3}{*}{SD3--Qwen}
& 1  & 31.375 & 0.9417 & 0.04028 \\
& 5  & 22.977 & 0.8431 & 0.12630 \\
& 10 & 20.043 & 0.7670 & 0.18892 \\

\midrule

\multirow{3}{*}{FLUX.1--Qwen}
& 1  & 32.579 & 0.9496 & 0.03663 \\
& 5  & 22.663 & 0.8450 & 0.13064 \\
& 10 & 19.106 & 0.7621 & 0.21014 \\

\midrule

\multirow{3}{*}{SD3--FLUX.1--Qwen}
& 1  & 32.397 & 0.9511 & 0.03257 \\
& 5  & 23.445 & 0.8580 & 0.11006 \\
& 10 & 20.119 & 0.7808 & 0.17591 \\

\bottomrule
\end{tabular}
}
}

\end{minipage}
\hfill
\begin{minipage}[t]{0.48\textwidth}
\centering

\subcaptionbox{
    Cross-architecture on-policy distillation.
    \label{tab:sd3_qwen_opd}
}[\linewidth]{
\resizebox{\linewidth}{!}{
\begin{tabular}{lccccc}
\toprule
\multirow{2}{*}{\textbf{Models}}
& \multicolumn{5}{c}{\textbf{HPSv2}} \\
\cmidrule(lr){2-6}

& \textbf{Anime}
& \textbf{Concept}
& \textbf{Paint}
& \textbf{Photo}
& \textbf{Avg.} \\
\midrule

Qwen-Image (Teacher)
& 29.16
& 28.22
& 28.40
& 25.90
& 27.92 \\

\midrule

SD3 Baseline
& \textbf{28.21}
& 26.25
& 26.01
& 24.59
& 26.27 \\

\rowcolor[HTML]{E2F4E3}
\textbf{SD3--Qwen OPD}
& 27.58
& \textbf{26.54}
& \textbf{26.98}
& \textbf{24.72}
& \textbf{26.46} \\

\bottomrule
\end{tabular}
}
}

\subcaptionbox{
    Reconstruction fidelity without $\mathcal{L}_{\mathrm{rec}}$.
    \label{tab:domain_reconstruction}
}[\linewidth]{
\resizebox{\linewidth}{!}{
\begin{tabular}{lccc}
\toprule
\textbf{Domain}
& \textbf{PSNR} $\uparrow$
& \textbf{SSIM} $\uparrow$
& \textbf{LPIPS} $\downarrow$ \\
\midrule

SD3 Average
& 10.4743
& 0.3627
& 0.9491 \\

Qwen-Image Average
& 8.6137
& 0.4794
& 0.9328 \\

\bottomrule
\end{tabular}
}
}

\end{minipage}

\end{table}

\begin{table}[t]
    \centering
    \caption{Comparison of base models and cross-model combinations, demonstrating that guidance from a small set of high-quality generators can improve the generation quality of weaker models.}
    \label{tab:backbone_comparison}

    \setlength{\tabcolsep}{5.5pt}
    \renewcommand{\arraystretch}{1.08}

    \resizebox{1\textwidth}{!}{
    \begin{tabular}{l ccc c ccccccc}
        \toprule

        \multirow{2}{*}{\textbf{Models}}
        & \multirow{2}{*}{\shortstack{\textbf{New}\\\textbf{Model}}}
        & \multirow{2}{*}{\shortstack{\textbf{GenEval-v2}}}
        & \multirow{2}{*}{\shortstack{\textbf{CVTG-2K}}}
        & \multirow{2}{*}{\shortstack{\textbf{Clip}}}
        & \multicolumn{6}{c}{\textbf{OneIG-Bench}}
        \\

        \cmidrule(lr){6-11}

        & & & & &
        \textbf{Alignment}
        & \textbf{Text}
        & \textbf{Diversity}
        & \textbf{Style}
        & \textbf{Reasoning}
        & \textbf{Avg.}
        \\

        \midrule

        \rowcolor{gray!18}
        \multicolumn{11}{l}{
            $\blacktriangledown$ \emph{Stronger Model}
        }
        \\

        Qwen-Image~\citep{wu2025qwen}
        & --
        & \textbf{41.73}
        & 0.6699
        & 25.46
        & 0.8304
        & 0.9138
        & 0.2091
        & 0.2230
        & 0.2486
        & 0.4849
        \\

        \midrule

        \rowcolor{gray!18}
        \multicolumn{11}{l}{
            $\blacktriangledown$ \emph{Generation Baselines}
        }
        \\

        SD3~\citep{esser2024scaling}
        & --
        & 22.14
        & 0.3682
        & 28.65
        & 0.7934
        & 0.4161
        & 0.2327
        & 0.3596
        & 0.2770
        & 0.4157
        \\

        FLUX.1~\citep{blackforestlabs2025flux2}
        & --
        & 22.86
        & 0.5180
        & 27.75
        & 0.7894
        & 0.5736
        & 0.2136
        & 0.3663
        & 0.2653
        & 0.4416
        \\

        \midrule

        \rowcolor{gray!18}
        \multicolumn{11}{l}{
            $\blacktriangledown$ \emph{Hybrid Models (Ours)}
        }
        \\

        
        \rowcolor[HTML]{E2F4E3}
        \textbf{Qwen-Image + SD3}
        & $\checkmark$
        & \scoregain{35.06}{12.92}
        & \scoregain{0.6488}{0.2806}
        & \scoreonly{26.98}
        & \scoregain{0.8072}{0.0138}
        & \scoregain{0.6683}{0.2522}
        & \scoregain{0.2645}{0.0318}
        & \scoregain{0.3748}{0.0152}
        & \scoreonly{0.2730}
        & \scoregain{0.4776}{0.0619}
        \\
        
        \rowcolor[HTML]{E2F4E3}
        \textbf{Qwen-Image + FLUX.1}
        & $\checkmark$
        & \scoregain{36.54}{13.68}
        & \scoregain{0.6983}{0.1803}
        & \scoregain{28.86}{1.11}
        & \scoregain{0.8262}{0.0368}
        & \scoregain{0.7854}{0.2118}
        & \scoregain{0.2269}{0.0133}
        & \scoregain{0.3785}{0.0122}
        & \scoregain{0.2868}{0.0215}
        & \scoregain{0.5007}{0.0591}
        \\
        
        \arrayrulecolor{black}\bottomrule
    \end{tabular}
    }
\end{table}

\begin{table}[H]
    \centering
    \caption{\textbf{Quantitative comparison of generation quality and inference efficiency.} Performance benchmarks across baselines, existing acceleration techniques, and InstantFusion. Furthermore, InstantFusion exhibits orthogonal compatibility when integrated with training-free acceleration methods like DiCache, yielding cumulative speedups without compromising generation fidelity.}
    \label{tab:main_results}

    \setlength{\tabcolsep}{3pt}
    \renewcommand{\arraystretch}{1.08}

    \resizebox{\textwidth}{!}{
    \begin{tabular}{lccccccccccc}
        \toprule

        \multirow{2}{*}{\textbf{Models}}
        & \multirow{2}{*}{\shortstack{\textbf{New}\\\textbf{model}}}
        & \multirow{2}{*}{\shortstack{\textbf{Time}\\$\downarrow$}}
        & \multirow{2}{*}{\shortstack{\textbf{GenEval}\\\textbf{-v2} $\uparrow$}}
        & \multirow{2}{*}{\shortstack{\textbf{Clip}\\$\uparrow$}}
        & \multirow{2}{*}{\shortstack{\textbf{Aesthetic}\\$\uparrow$}}
        & \multicolumn{6}{c}{\textbf{OneIG-Bench} $\uparrow$}
        \\

        \cmidrule(lr){7-12}

        &
        &
        &
        &
        &
        &
        \textbf{Alignment}
        &
        \textbf{Text}
        &
        \textbf{Diversity}
        &
        \textbf{Style}
        &
        \textbf{Reasoning}
        &
        \textbf{Avg.}
        \\

        \midrule

        \rowcolor{gray!18}
        \multicolumn{12}{l}{
            $\blacktriangledown$ \emph{Generation Baselines}
        }
        \\

        Qwen-Image~\citep{wu2025qwen}
        & --
        & 18.84
        & 41.73
        & 25.46
        & 5.9797
        & 0.8304
        & 0.9138
        & 0.2091
        & 0.2230
        & 0.2486
        & 0.4849
        \\

        Qwen-Few-Step
        & --
        & 0.893
        & 32.94
        & 27.48
        & 6.1430
        & 0.8804
        & 0.9752
        & 0.0885
        & 0.3234
        & 0.3069
        & 0.5148
        \\

        FLUX.2-Klein~\citep{blackforestlabs2025flux2}
        & --
        & 0.516
        & 25.43
        & 27.71
        & 5.9017
        & 0.5718
        & 0.1863
        & \textbf{0.3930}
        & 0.4059
        & 0.2415
        & 0.3597
        \\

        Z-Image-Turbo~\citep{cai2025z}
        & --
        & 3.521
        & 39.21
        & 25.02
        & 4.9892
        & 0.8658
        & 0.9788
        & 0.2193
        & 0.4573
        & \textbf{0.3137}
        & \textbf{0.5669}
        \\

        T-Stitch~\citep{pan2025t}
        & --
        & 20.76
        & 7.762
        & 21.24
        & 4.4475
        & 0.3037
        & 0.3349
        & 0.2445
        & 0.2255
        & 0.2208
        & 0.2659
        \\

        \midrule

        \rowcolor{gray!18}
        \multicolumn{12}{l}{
            $\blacktriangledown$ \emph{InstantFusion (Ours)}
        }
        \\

        \rowcolor[HTML]{E2F4E3}
        \textbf{Z-Image-Turbo + Qwen-Image}
        & $\checkmark$
        & \underline{4.373}
        & \textbf{33.34}
        & 27.90
        & 5.3957
        & 0.8043
        & 0.8837 
        & \underline{0.3070}
        & \underline{0.3933}
        & 0.3046
        & 0.5386
        \\

        \rowcolor[HTML]{E2F4E3}
        \textbf{FLUX.2-Klein + Qwen-Image}
        & $\checkmark$
        & \textbf{2.545}
        & 27.19
        & 29.18
        & 5.7441
        & 0.8444
        & 0.9371
        & 0.2454
        & \textbf{0.3977}
        & \underline{0.3089}
        & \underline{0.5467}
        \\

        \rowcolor[HTML]{E2F4E3}
        \textbf{Qwen-Few-Step + Qwen-Image}
        & $\checkmark$
        & 10.64
        & \underline{32.70}
        & \underline{33.15}
        & \underline{6.0821}
        & \textbf{0.8808}
        & \underline{0.9800}
        & 0.1192
        & 0.3205
        & 0.3015
        & 0.5204
        \\

        \rowcolor[HTML]{E2F4E3}
        \textbf{Qwen-Few-Step + Qwen-Image + DiCache}
        & $\checkmark$
        & 5.700
        & 32.46
        & \textbf{33.18}
        & \textbf{6.1076}
        & \underline{0.8780}
        & \textbf{0.9812}
        & 0.1190
        & 0.3284
        & 0.3033
        & 0.5220
        \\

        \bottomrule
    \end{tabular}
        }
\end{table}




\begin{table}[t]
    \centering
    \caption{Evaluation results of reward-specialized GRPO models and model combination strategies.
    }
    \label{tab:reward_model_fusion}

    \setlength{\tabcolsep}{4.0pt}
    \renewcommand{\arraystretch}{1.08}

    \resizebox{\textwidth}{!}{
    \begin{tabular}{
        l
        cc
        ccccc
        ccccc
    }
        \toprule

        \multirow{2}{*}{\textbf{Models}}
        & \multirow{2}{*}{\shortstack{\textbf{New}\\\textbf{Model}}}
        & \multirow{2}{*}{\shortstack{\textbf{GenEval-v2}}}
        & \multicolumn{5}{c}{\textbf{HPSv2}}
        & \multicolumn{5}{c}{\textbf{HPSv3}}
        \\

        \cmidrule(lr){4-8}
        \cmidrule(lr){9-13}

        & & &
        \textbf{Anime}
        & \textbf{Concept}
        & \textbf{Paint.}
        & \textbf{Photo}
        & \textbf{Avg.}
        & \textbf{Anime}
        & \textbf{Concept}
        & \textbf{Paint.}
        & \textbf{Photo}
        & \textbf{Avg.}
        \\

        \midrule

        \rowcolor{gray!18}
        \multicolumn{13}{l}{
            $\blacktriangledown$ \emph{Reward-Specialized GRPO Baselines}
        }
        \\

        GRPO baseline (HPSv2)
        & --
        & 28.83
        & 32.21
        & 31.67
        & 31.91
        & 28.95
        & 31.18
        & 12.14
        & 12.88
        & 13.30
        & 9.94
        & 12.07
        \\

        GRPO baseline (HPSv3)
        & --
        & 26.53
        & 31.42
        & 30.57
        & 30.93
        & 27.63
        & 30.14
        & 12.31
        & 13.06
        & 13.48
        & 9.73
        & 12.15
        \\

        GRPO baseline (GenEval-v2)
        & --
        & 35.89
        & 28.55
        & 27.28
        & 27.47
        & 25.47
        & 27.19
        & 10.98
        & 11.30
        & 11.71
        & 7.69
        & 10.42
        \\

        \midrule

        \rowcolor{gray!18}
        \multicolumn{13}{l}{
            $\blacktriangledown$ \emph{Direct V Fusion}
        }
        \\

        HPSv2 + GenEval-v2
        & --
        & 29.39
        & 31.10
        & 30.58
        & 30.86
        & 27.40
        & 29.99
        & 11.62
        & 12.34
        & 12.78
        & 8.94
        & 11.42
        \\

        HPSv2 + HPSv3
        & --
        & 28.58
        & 32.11
        & 31.50
        & 31.72
        & 28.75
        & 31.02
        & 12.33
        & 13.03 
        & 13.46
        & 10.22
        & 12.26
        \\

        \rowcolor{gray!18}
        \multicolumn{13}{l}{
            $\blacktriangledown$ \emph{Multi-preference}
        }
        \\

        HPSv2 + GenEval-v2
        & --
        & \underline{30.49}
        & \underline{31.86}
        & 30.89
        & 31.22
        & \underline{28.41}
        & \underline{30.60}
        & 11.54
        & 11.92
        & 12.36
        & 9.55
        & 11.34
        \\

        HPSv2 + HPSv3
        & --
        & 24.53
        & 28.83
        & 27.36
        & 27.57
        & 25.70
        & 27.37
        & 11.58
        & 11.99
        & 12.34
        & 8.04
        & 10.99
        \\

        \rowcolor{gray!18}
        \multicolumn{13}{l}{
            $\blacktriangledown$ \emph{InstantFusion (Ours)}
        }
        \\

        \rowcolor[HTML]{E2F4E3}
        \textbf{GenEval-v2 + HPSv2}
        & $\checkmark$
        & \textbf{30.97}
        & 31.91
        & \underline{31.37}
        & \underline{31.52}
        & 27.92
        & 30.68
        & \underline{12.09}
        & \underline{12.72}
        & \underline{13.07}
        & \underline{9.61}
        & \underline{11.87}
        \\

        \rowcolor[HTML]{E2F4E3}
        \textbf{HPSv2 + HPSv3}
        & $\checkmark$
        & 28.94
        & \textbf{32.18}
        & \textbf{31.66}
        & \textbf{31.89}
        & \textbf{28.69}
        & \textbf{31.11}
        & \textbf{12.26}
        & \textbf{13.00}
        & \textbf{13.36}
        & \textbf{10.13}
        & \textbf{12.19}
        \\

        \bottomrule
    \end{tabular}
    }
\end{table}

\begin{table*}[t]
    \centering

    \begin{minipage}[t]{0.48\textwidth}
        \centering
        \captionof{table}{Capabilities enabled by progressively adding the training objectives. Each column indicates whether the resulting interface supports the corresponding capability.}
        \label{tab:objective_capabilities}

        \setlength{\tabcolsep}{4pt}
        \renewcommand{\arraystretch}{1.08}

        \resizebox{\linewidth}{!}{
        \begin{tabular}{lccc}
            \toprule
            \textbf{Training Objectives}
            & \shortstack{\textbf{Cross-Model}\\\textbf{State Translation}}
            & \shortstack{\textbf{Information-Preserving}\\\textbf{Reconstruction}}
            & \shortstack{\textbf{Denoising-Dynamics}\\\textbf{Consistency}}
            \\
            \midrule

            $\mathcal{L}_{\mathrm{align}}$
            & $\checkmark$
            & $\times$
            & $\times$
            \\

            $\mathcal{L}_{\mathrm{align}}
            +\mathcal{L}_{\mathrm{rec}}$
            & $\checkmark$
            & $\checkmark$
            & $\times$
            \\

            \textbf{InstantFusion}
            & $\checkmark$
            & $\checkmark$
            & $\checkmark$
            \\

            \bottomrule
        \end{tabular}
        }
    \end{minipage}
    \hfill
    \begin{minipage}[t]{0.48\textwidth}
        \centering
        \captionof{table}{Comparison between the baseline and InstantFusion for cross-model transfer. InstantFusion consistently outperforms the baseline across all evaluated metrics}
        \label{tab:lae_architecture}

        \setlength{\tabcolsep}{4pt}
        \renewcommand{\arraystretch}{1.08}

        \resizebox{\linewidth}{!}{
        \begin{tabular}{llccc}
            \toprule
            \textbf{Transfer Direction}
            & \textbf{Method}
            & \shortstack{\textbf{GenEval-v2}}
            & \shortstack{\textbf{CVTG-2K}}
            & \shortstack{\textbf{HPSv2 Avg.}}
            \\
            \midrule

            \multirow{2}{*}{Qwen-Image $\rightarrow$ SD3}
            & Conv. layer baseline
            & 27.86
            & 0.2744
            & 20.50
            \\

            & \textbf{InstantFusion}
            & 35.06
            & 0.6488
            & 25.34
            \\

            \midrule

            \multirow{2}{*}{Qwen-Image $\rightarrow$ FLUX.1}
            & Conv. layer baseline
            & 17.37
            & 0.3566
            & 14.79
            \\

            & \textbf{InstantFusion}
            & 36.54
            & 0.6983
            & 25.03
            \\

            \bottomrule
        \end{tabular}
        }
    \end{minipage}

\end{table*}


\subsection{Applications of LAE}
\label{sec:what}

\label{sec:applications}



\textbf{Generation Acceleration}.
Tab.~\ref{tab:main_results} shows that InstantFusion accelerates
Qwen-Image by combining it with efficient generators. Pairing Qwen-Image
with Z-Image-Turbo or FLUX.2-Klein reduces inference time from $18.84$s
to $4.373$s and $2.545$s, while achieving GenEval-v2
scores of $33.34$, $27.19$ and OneIG-Bench averages of $0.5386$, $0.5467$, without the quality collapse observed in T-Stitch~\citep{pan2025t}.  InstantFusion is further compatible with existing acceleration methods: applying DiCache~\citep{bu2026dicache} to
InstantFusion reduces latency from $10.64$s to $5.70$s,
while GenEval-v2 changes only from $32.70$ to $32.46$ and the
OneIG-Bench average marginally increases.

\begin{figure}[t]
    \centering
    \includegraphics[width=1\linewidth]{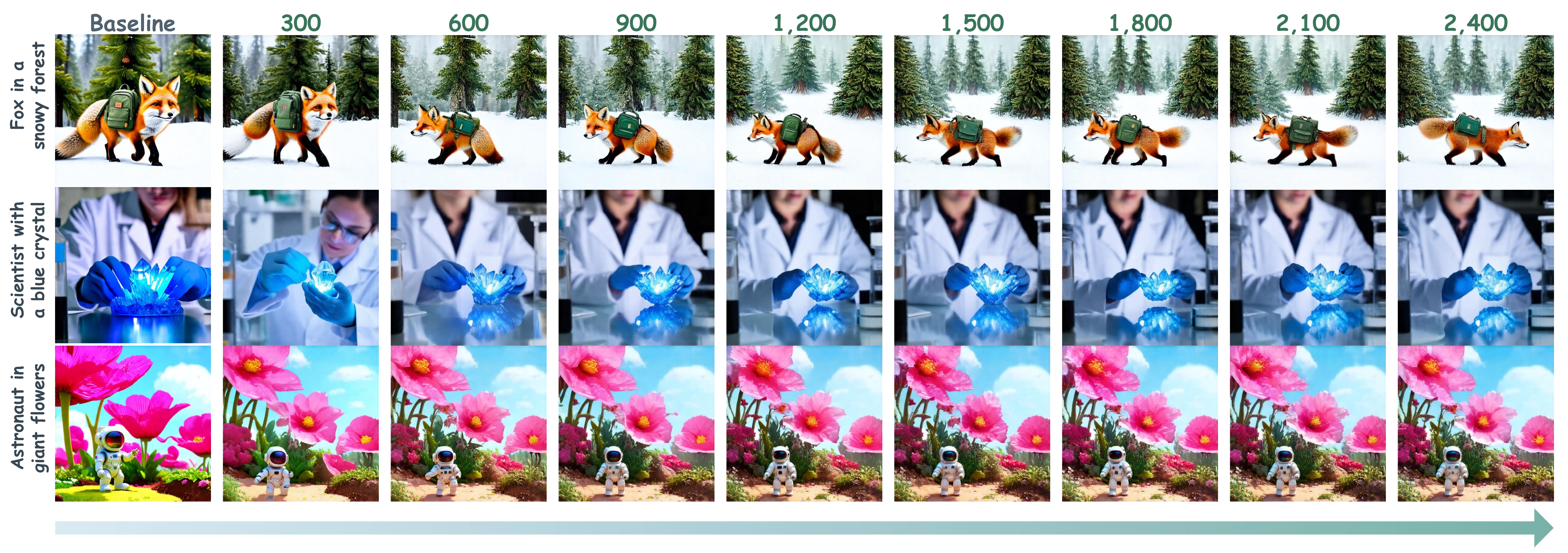}
    \caption{Evolution of generated images over training steps during cross-model OPD.}
    \label{fig:cross_model_opd}
\end{figure}

\textbf{Cross-Architecture On-Policy Distillation}. We leverage InstantFusion for cross-architecture on-policy distillation (OPD), transferring generative priors from a stronger teacher (Qwen-Image) to a student (SD3)~\citep{jiang2026dopsd,li2026diffusionopd,fang2026flow,zhou2026danceopd,fu2026poly,yuan2026vision}. As shown in Fig.~\ref{fig:method_overview}(b), InstantFusion projects intermediate states from both models into the shared latent space, enabling the teacher to compute velocity targets directly along the student's trajectory. Consequently, Tab.~\ref{tab:sd3_qwen_opd} shows that cross-model OPD improves SD3 from 26.27 to 26.46 in HPSv2, with qualitative evolutions illustrated in Fig.~\ref{fig:cross_model_opd}.

\textbf{Multi-Preference Composition}. Tab.~\ref{tab:reward_model_fusion} evaluates the composition of independently trained reward experts~\citep{lee2025calibrated,rame2023rewarded}. When combining experts optimized for GenEval-v2 and HPSv2, InstantFusion achieves scores of $30.97$ and $30.68$, outperforming velocity fusion and closely matching jointly trained multi-reward GRPO. This result shows that InstantFusion can combine existing single-reward experts without the additional optimization required by joint multi-reward training. We observe the same trend when composing HPSv2 and HPSv3 experts, indicating that the shared latent space preserves the respective preferences learned by the individual experts.

\subsection{Ablation and Analysis}
\label{sec:ablation}

\textbf{Ablation of Training Objectives.} The three objectives serve complementary functions: $\mathcal{L}_{\mathrm{align}}$ enables cross-model state translation, $\mathcal{L}_{\mathrm{rec}}$ preserves information fidelity for accurate reconstruction, and $\mathcal{L}_{\mathrm{dyn}}$ aligns denoising trajectories across architectures (Tab.~\ref{tab:objective_capabilities}). As shown in Fig.~\ref{fig:dynamics_alignment_failure}, omitting $\mathcal{L}_{\mathrm{dyn}}$ causes training divergence during cross-architecture OPD, even with aligned latent spaces. Conversely, removing $\mathcal{L}_{\mathrm{rec}}$ allows cross-model translation but significantly impairs reconstruction quality. These distinct failure modes demonstrate that all three loss terms are essential.



\begin{figure}[t]
    \centering
    \includegraphics[width=1\linewidth]{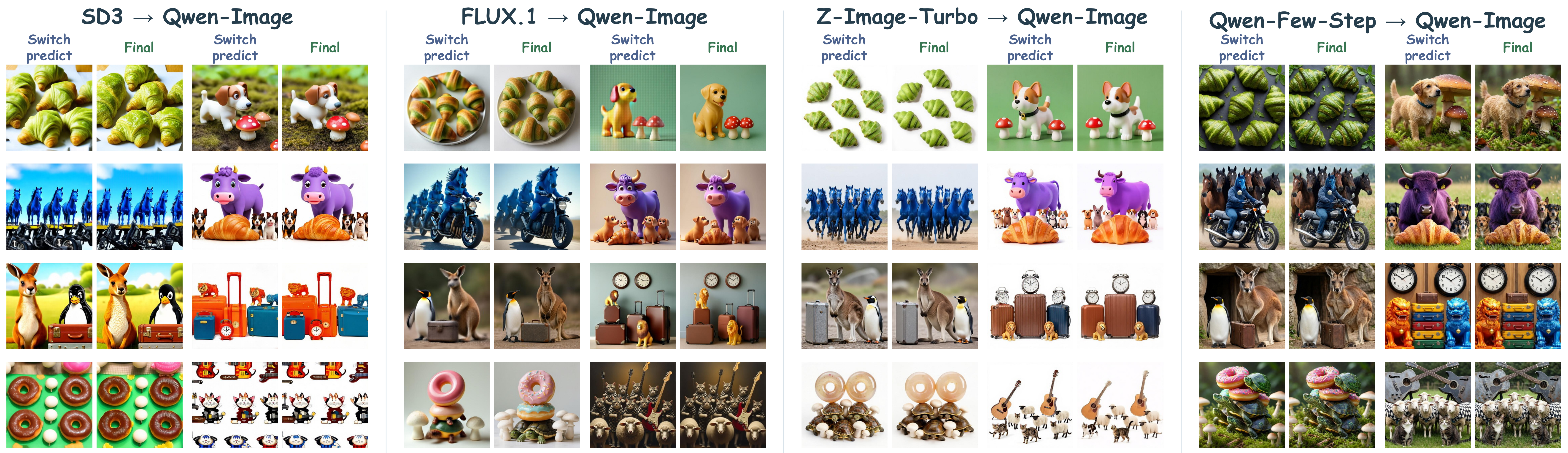}
    \caption{Comparison between the predicted clean images at adaptively selected handoff steps and the corresponding fully denoised outputs, demonstrating preservation of global structure.}
    \label{fig:adaptive_handoff}
\end{figure}

\textbf{Adaptive Handoff Analysis.} To verify whether the adaptive criterion selects a suitable transition step, we compare the clean image predicted at the selected handoff step with the final generated image. As shown in Fig.~\ref{fig:adaptive_handoff}, this early prediction already preserves the coarse spatial layout of the final output. Crucially, this structural consistency indicates that the selected handoff often occurs after the global structure has largely stabilized.

\textbf{LAE Architecture Design.} We conduct an ablation to validate the effectiveness of our time-conditioned residual blocks in LAE. Replacing these time-conditioned blocks with standard convolutional layers leads to a substantial drop in cross-model handoff performance (Tab.~\ref{tab:lae_architecture}), reducing GenEval-v2 by $7.20$ when combining Qwen-Image with SD3 and by $19.17$ with FLUX.1. This performance gap under an identical layer highlights the representational efficiency of our design.

\subsection{Ablation on Three-Stage Rollout Scheduling}

To verify whether the shared latent interface remains stable under multi-stage model cascading, we evaluate all six permutation schedules among SD3, FLUX.1, and Qwen-Image across a three-stage generation pipeline (Tab.~\ref{tab:three_stage_scheduling}). Across GenEval-v2 and HPSv2 metrics, the generation quality remains highly consistent regardless of the specific model execution order.

\begin{table*}[t]
    \centering
    \caption{
    \textbf{Robustness of the shared latent interface across multi-stage rollout schedules.} We evaluate different execution orders across three heterogeneous generators (SD3, FLUX.1, and Qwen-Image). The consistent performance across arbitrary transition permutations confirms the stability and invariance of the joint shared latent space under cascading cross-model transfers.
    }
    \label{tab:three_stage_scheduling}

    \setlength{\tabcolsep}{5pt}
    \renewcommand{\arraystretch}{1.08}

    \resizebox{\textwidth}{!}{
    \begin{tabular}{lcccccc}
        \toprule
        \multirow{2}{*}{\textbf{Rollout Schedule}}
        & \multirow{2}{*}{
            \shortstack{\textbf{GenEval-v2}}
        }
        & \multicolumn{5}{c}{\textbf{HPSv2}} \\
        \cmidrule(lr){3-7}

        &
        & \textbf{Anime}
        & \textbf{Concept}
        & \textbf{Paint.}
        & \textbf{Photo}
        & \textbf{Avg.} \\
        \midrule

        SD3 $\rightarrow$ FLUX.1 $\rightarrow$ Qwen
        & 19.99 & 27.04 & 25.41 & 25.46 & 24.69 & 25.65 \\

        SD3 $\rightarrow$ Qwen $\rightarrow$ FLUX.1
        & 24.93 & 24.35 & 22.80 & 22.72 & 21.91 & 22.95 \\

        FLUX.1 $\rightarrow$ SD3 $\rightarrow$ Qwen
        & 25.22 & 27.86 & 25.41 & 25.30 & 24.87 & 25.86 \\

        FLUX.1 $\rightarrow$ Qwen $\rightarrow$ SD3
        & 27.96 & 25.68 & 22.85 & 22.96 & 22.68 & 23.54 \\

        Qwen $\rightarrow$ SD3 $\rightarrow$ FLUX.1
        & 30.76 & 26.60 & 24.14 & 24.23 & 24.06 & 24.76 \\

        Qwen $\rightarrow$ FLUX.1 $\rightarrow$ SD3
        & 33.13 & 27.11 & 24.62 & 24.98 & 23.71 & 25.11 \\

        \bottomrule
    \end{tabular}
    }
\end{table*}

\subsection{Inference Efficiency Analysis}

To determine whether the shared latent interface introduces measurable
inference overhead, we compare the end-to-end latency of individual
generators and LAE-enabled hybrid generation. Two-model hybrids allocate the steps
evenly, while the three-model hybrid uses a $17/17/16$
allocation. After three warm-up runs, each configuration is
measured over 20 runs, with the results reported in
Table~\ref{tab:inference_latency}. Since each handoff only translates the latent state without
invoking an additional denoiser evaluation, InstantFusion introduces no
extra theoretical NFE.

\begin{table*}[t]
    \centering
    \caption{\small
    \textbf{End-to-end inference latency and theoretical NFEs.}
    Latency is measured at $1024\times1024$ resolution, averaged over
    20 runs, and reported in seconds per image. For hybrid configurations,
    the NFE entries show how the 50 denoising evaluations are allocated
    across models.
    }
    \label{tab:inference_latency}

    \setlength{\tabcolsep}{4pt}
    \renewcommand{\arraystretch}{1.08}

    \resizebox{\textwidth}{!}{
    \begin{tabular}{lccccccc}
        \toprule
        \multirow{2}{*}{\textbf{Metric}}
        & \multicolumn{3}{c}{\textbf{Individual Generators}}
        & \multicolumn{4}{c}{\textbf{Hybrid Generation}} \\
        \cmidrule(lr){2-4}
        \cmidrule(lr){5-8}

        & \textbf{SD3}
        & \textbf{FLUX.1}
        & \textbf{Qwen}
        & \textbf{FLUX.1$\rightarrow$SD3}
        & \textbf{SD3$\rightarrow$Qwen}
        & \textbf{FLUX.1$\rightarrow$Qwen}
        & \textbf{FLUX.1$\rightarrow$SD3$\rightarrow$Qwen} \\
        \midrule

        \textbf{Latency (s/image)}
        & $3.61$
        & $10.70$
        & $20.82$
        & $7.16$
        & $12.21$
        & $15.7$
        & $11.53$ \\

        \textbf{Theoretical NFE}
        & $50$
        & $50$
        & $50$
        & $25+25$
        & $25+25$
        & $25+25$
        & $17+17+16$ \\

        \bottomrule
    \end{tabular}
    }
\end{table*}

\section{Limitations}
\label{app:limitations}

While InstantFusion provides a lightweight interface for cross-architecture state and dynamics translation without pair-specific mappings, several limitations remain. Currently, our formulation relies on a shared autoencoder and compatible time parameterization across generators. Furthermore, executing hybrid inference requires hosting multiple frozen backbones simultaneously, incurring elevated memory and storage overhead. Extending this interface across heterogeneous autoencoders and optimizing dynamic model loading/offloading represent key directions for future research.

\section{Conclusion}
\vspace{-0.5em}

In this work, we introduced \textbf{InstantFusion}, a shared latent interface that connects heterogeneous image generators by aligning their latent states. Without modifying the frozen generators, InstantFusion enables cross-model state transfer. Results across a diverse range of downstream tasks further demonstrate the flexibility and broad applicability of InstantFusion. Future work may develop the learned shared space into a reusable pretrained interface, such that new generators can be incorporated with minimal adaptation and benefit from a common starting point for model composition, knowledge transfer, and downstream optimization.

\clearpage
\bibliographystyle{plain}
\bibliography{refs}


\clearpage
\appendix
\section{Appendix}
The supplementary material is organized as follows.
Sec.~\ref{app:details} provides implementation and training details
for the shared latent interface and cross-model OPD;
Sec.~\ref{app:relatedwork} reviews additional related work;
Sec.~\ref{app:quantitative} reports extended quantitative results,
including additional benchmarks, efficiency analysis, and rollout
scheduling ablations;
Sec.~\ref{app:qualitative} presents additional qualitative examples
and handoffs at different denoising steps; and
Sec.~\ref{app:limitations} discusses the limitations of our method.

\section{Training details}
\label{app:details}

Tables~\ref{tab:lae_training_config} and
\ref{tab:opd_training_config} summarize the training configurations.
The first stage optimizes only the model-specific LAEs, while
cross-model OPD freezes the teacher, reference model, and latent
interface and updates only the student LoRA.

\begin{table*}[h]
\centering

\begin{minipage}[t]{0.485\textwidth}
    \centering
    \caption{
    \textbf{Training hyperparameters for the shared latent interface.}
    }
    \label{tab:lae_training_config}

    \small
    \setlength{\tabcolsep}{3pt}
    \renewcommand{\arraystretch}{1.08}

    \begin{tabular}{
        @{}l
        >{\raggedright\arraybackslash}p{0.55\linewidth}@{}
    }
        \toprule
        \textbf{Hyperparameter} & \textbf{Setting} \\
        \midrule

        Architecture
        & ResBlock LAE \\

        Latent channels
        & 16 native; 32 shared \\

        Hidden channels
        & 64 \\

        ResBlocks depth 
        & 3 \\

        $\sigma$ embedding size
        & 128 \\

        Trainable parameters
        & 1.532M \\

        Optimizer
        & AdamW \\

        Learning rate
        & $10^{-4}$ \\

        Weight decay
        & $0.01$ \\
        
        Batch size
        & 1 per GPU; accumulation 4;
          effective batch size 32 \\

        Precision
        & BF16 \\

        GPUs
        & 8 \\

        Time sampling
        & 1,000 timesteps;
          exponential shift $\mu=\ln 3$ \\

        Optimization
        & LAEs only \\

        \bottomrule
    \end{tabular}
\end{minipage}
\hfill
\begin{minipage}[t]{0.485\textwidth}
    \centering
    \caption{
    \textbf{Training hyperparameters for cross-model OPD.}
    }
    \label{tab:opd_training_config}

    \small
    \setlength{\tabcolsep}{3pt}
    \renewcommand{\arraystretch}{1.08}

    \begin{tabular}{
        @{}l
        >{\raggedright\arraybackslash}p{0.55\linewidth}@{}
    }
        \toprule
        \textbf{Hyperparameter} & \textbf{Setting} \\
        \midrule

        Sampling steps
        & 20 \\

        Supervision steps
        & 1, 2, 4, 8 \\

        Optimizer
        & AdamW \\

        Learning rate
        & $10^{-5}$ \\

        Weight decay
        & $0.01$ \\

        Batch size
        & 1 per GPU; accumulation 2;
          effective batch size 16 \\

        Precision 
        & BF16 \\

        GPUs
        & 8 \\

        LoRA configuration
        & Rank 64; alpha 64;
          Kaiming initialization \\

        LoRA targets
        & Attention QKV/output and
          FFN projections \\

        Reference loss weight
        & $0.1$ \\

        Training length
        & Up to 50 epochs \\

        Optimization
        & SD3 LoRA only \\

        \bottomrule
    \end{tabular}
\end{minipage}

\end{table*}

\section{Additional Related Work}
\label{app:relatedwork}

\textbf{Efficient Generation}. Existing acceleration methods reduce either the number or the computational cost of denoising evaluations. Few-step methods compress long denoising trajectories into one or a few evaluations through distillation or consistency learning~\citep{salimans2022progressive, song2023consistency, yin2024one, luo2023latent}, while caching-based methods reduce per-step computation by reusing or forecasting intermediate features across adjacent timesteps~\citep{liu2025timestep,bu2026dicache,liu2025reusing}. Model scheduling methods such as OMS-DPM and T-Stitch further assign different pretrained models to different denoising stages~\citep{liu2023oms,pan2025t}. These studies demonstrate the potential of multi-model acceleration, but have primarily considered models whose intermediate states are directly interchangeable. Composing independently developed generators with heterogeneous latent spaces remains largely unexplored.

\textbf{Diffusion On-Policy Distillation}. On-policy distillation (OPD) has proven effective in large language models~\citep{gu2023minillm,agarwal2024policy}. Its extensions to diffusion and flow models follow two main directions: on-policy self-distillation uses privileged information to construct a stronger teacher from the student itself~\citep{jiang2026dopsd}, while teacher-based OPD transfers denoising capabilities from stronger pretrained models along student trajectories~\citep{li2026diffusionopd,fang2026flow,zhou2026danceopd,yuan2026vision}. Most methods, however, either assume that teacher and student belong to the same model family or bridge heterogeneous models by comparing their decoded outputs in a frozen visual representation~\citep{fu2026any,fu2026poly}. How to directly transfer a teacher's denoising dynamics to a student with an incompatible latent space therefore remains an open question.

\textbf{Multi-Preference Composition}. Multi-preference generation is commonly formulated as learning a single model that balances several objectives. Prior methods either jointly optimize one generator with multiple reward signals or separately fine-tune preference-specific models and merge their parameters~\citep{lee2025calibrated,rame2023rewarded}. The former encodes preference trade-offs through multi-objective training, whereas the latter relies on compatible architectures and weight-space interpolation. In both cases, preference composition remains tied to model training or parameter manipulation, limiting the flexible reuse of independently trained experts.

\section{Per-Model Reconstruction Analysis Beyond Average Metrics}
\label{app:quantitative}

\label{abl:detailed_reconstruction}

While Table~\ref{tab:reconstruction} summarizes the average reconstruction fidelity across generators, Table~\ref{tab:full_reconstruction} provides a granular per-model breakdown—reporting PSNR, SSIM, and LPIPS across 1, 5, and 10 latent round trips. This detailed evaluation highlights how reconstruction errors accumulate over sequential cross-model translations, confirming the stability limits of the shared latent space.



\section{Extended Qualitative Results}
\label{app:qualitative}

\subsection{Handoffs at Different Denoising Steps}

\begin{table}[t]
    \centering
    \caption{Reconstruction fidelity under different shared-space configurations.
    }
    \label{tab:full_reconstruction}

    \setlength{\tabcolsep}{5pt}
    \renewcommand{\arraystretch}{1.08}

    \resizebox{\linewidth}{!}{
    \begin{tabular}{l l c ccc}
        \toprule

        \textbf{Shared Space}
        & \textbf{Reconstructed Model}
        & \textbf{Switch}
        & \textbf{PSNR} $\uparrow$
        & \textbf{SSIM} $\uparrow$
        & \textbf{LPIPS} $\downarrow$
        \\

        \midrule

       TaylorSeer ($\mathcal{N}=6$)
        & FLUX.1
        & --
        & 28.945
        & 0.6556
        & 0.4020
        \\

        \midrule

        \multirow{9}{*}{SD3--Qwen}
        & \multirow{3}{*}{Qwen}
        & 1
        & 32.418
        & 0.9433
        & 0.03811
        \\
        & & 5
        & 24.653
        & 0.8719
        & 0.10096
        \\
        & & 10
        & 22.255
        & 0.8253
        & 0.13709
        \\

        \cmidrule(lr){2-6}

        & \multirow{3}{*}{SD3}
        & 1
        & 30.331
        & 0.9401
        & 0.04244
        \\
        & & 5
        & 21.301
        & 0.8143
        & 0.15164
        \\
        & & 10
        & 17.830
        & 0.7086
        & 0.24075
        \\

        \cmidrule(lr){2-6}

        & \multirow{3}{*}{Average}
        & 1
        & 31.375
        & 0.9417
        & 0.04028
        \\
        & & 5
        & 22.977
        & 0.8431
        & 0.12630
        \\
        & & 10
        & 20.043
        & 0.7670
        & 0.18892
        \\

        \midrule

        \multirow{9}{*}{FLUX.1--Qwen}
        & \multirow{3}{*}{FLUX.1}
        & 1
        & 33.253
        & 0.9575
        & 0.02897
        \\
        & & 5
        & 21.512
        & 0.8326
        & 0.13814
        \\
        & & 10
        & 16.886
        & 0.7171
        & 0.24954
        \\

        \cmidrule(lr){2-6}

        & \multirow{3}{*}{Qwen}
        & 1
        & 31.904
        & 0.9416
        & 0.04428
        \\
        & & 5
        & 23.813
        & 0.8573
        & 0.12314
        \\
        & & 10
        & 21.325
        & 0.8070
        & 0.17074
        \\

        \cmidrule(lr){2-6}

        & \multirow{3}{*}{Average}
        & 1
        & 32.579
        & 0.9496
        & 0.03663
        \\
        & & 5
        & 22.663
        & 0.8450
        & 0.13064
        \\
        & & 10
        & 19.106
        & 0.7621
        & 0.21014
        \\

        \midrule

        \multirow{12}{*}{SD3--FLUX.1--Qwen}
        & \multirow{3}{*}{FLUX.1}
        & 1
        & 34.159
        & 0.9639
        & 0.02168
        \\
        & & 5
        & 23.799
        & 0.8657
        & 0.10111
        \\
        & & 10
        & 19.571
        & 0.7814
        & 0.17821
        \\

        \cmidrule(lr){2-6}

        & \multirow{3}{*}{Qwen}
        & 1
        & 32.276
        & 0.9441
        & 0.03814
        \\
        & & 5
        & 23.982
        & 0.8670
        & 0.10392
        \\
        & & 10
        & 21.459
        & 0.8191
        & 0.14158
        \\

        \cmidrule(lr){2-6}

        & \multirow{3}{*}{SD3}
        & 1
        & 30.757
        & 0.9453
        & 0.03788
        \\
        & & 5
        & 22.553
        & 0.8412
        & 0.12516
        \\
        & & 10
        & 19.328
        & 0.7418
        & 0.20794
        \\

        \cmidrule(lr){2-6}

        & \multirow{3}{*}{Average}
        & 1
        & 32.397
        & 0.9511
        & 0.03257
        \\
        & & 5
        & 23.445
        & 0.8580
        & 0.11006
        \\
        & & 10
        & 20.119
        & 0.7808
        & 0.17591
        \\

        \bottomrule
    \end{tabular}
    }
\end{table}

\begin{figure}[t]
    \centering
    \includegraphics[width=1\linewidth]{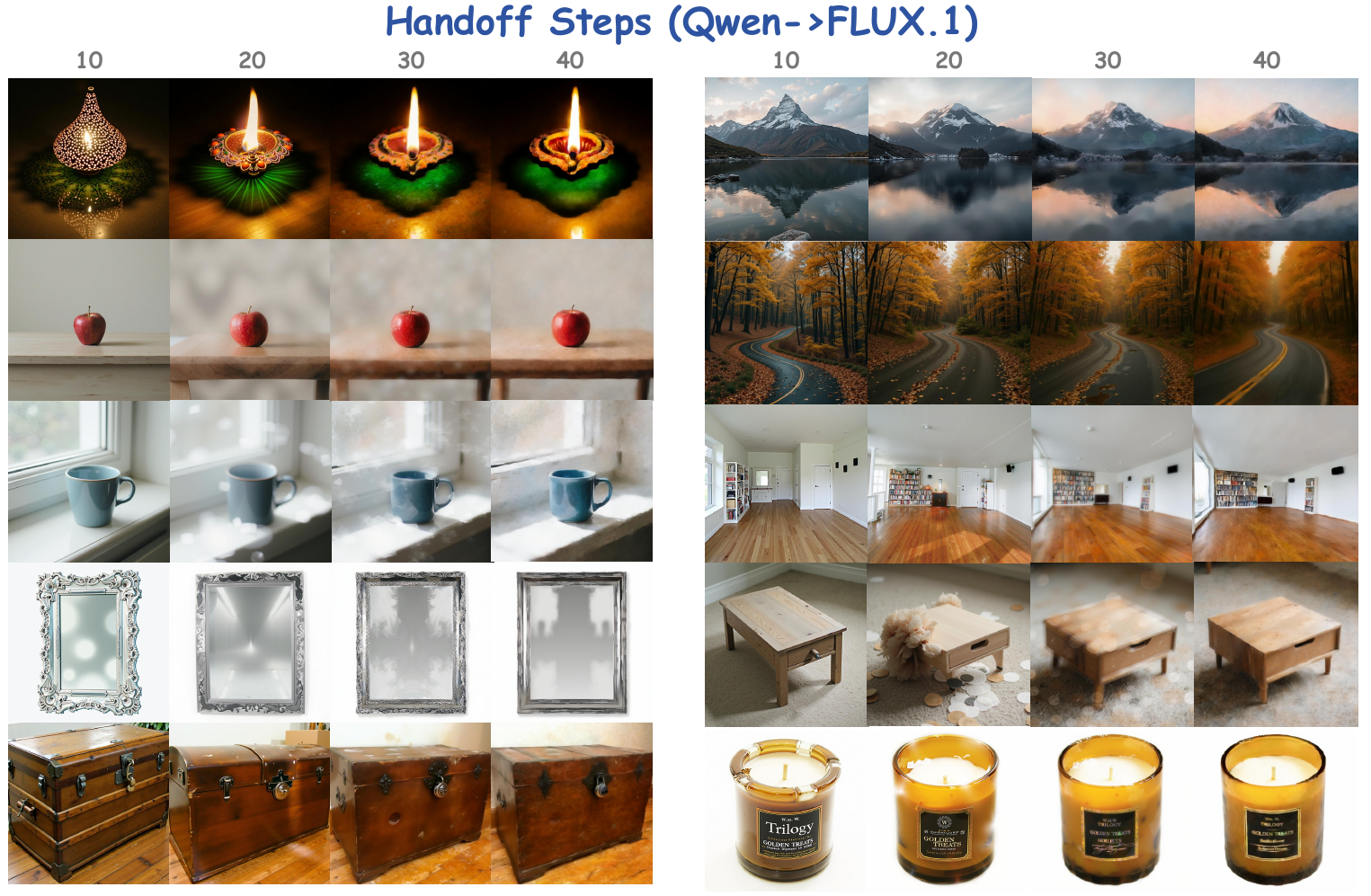}
    \caption{Cross-model state handoffs at different denoising timesteps (Qwen-Image, Flux.1).}
    \label{fig:app_switch_step_1}
\end{figure}

\begin{figure}[t]
    \centering
    \includegraphics[width=1\linewidth]{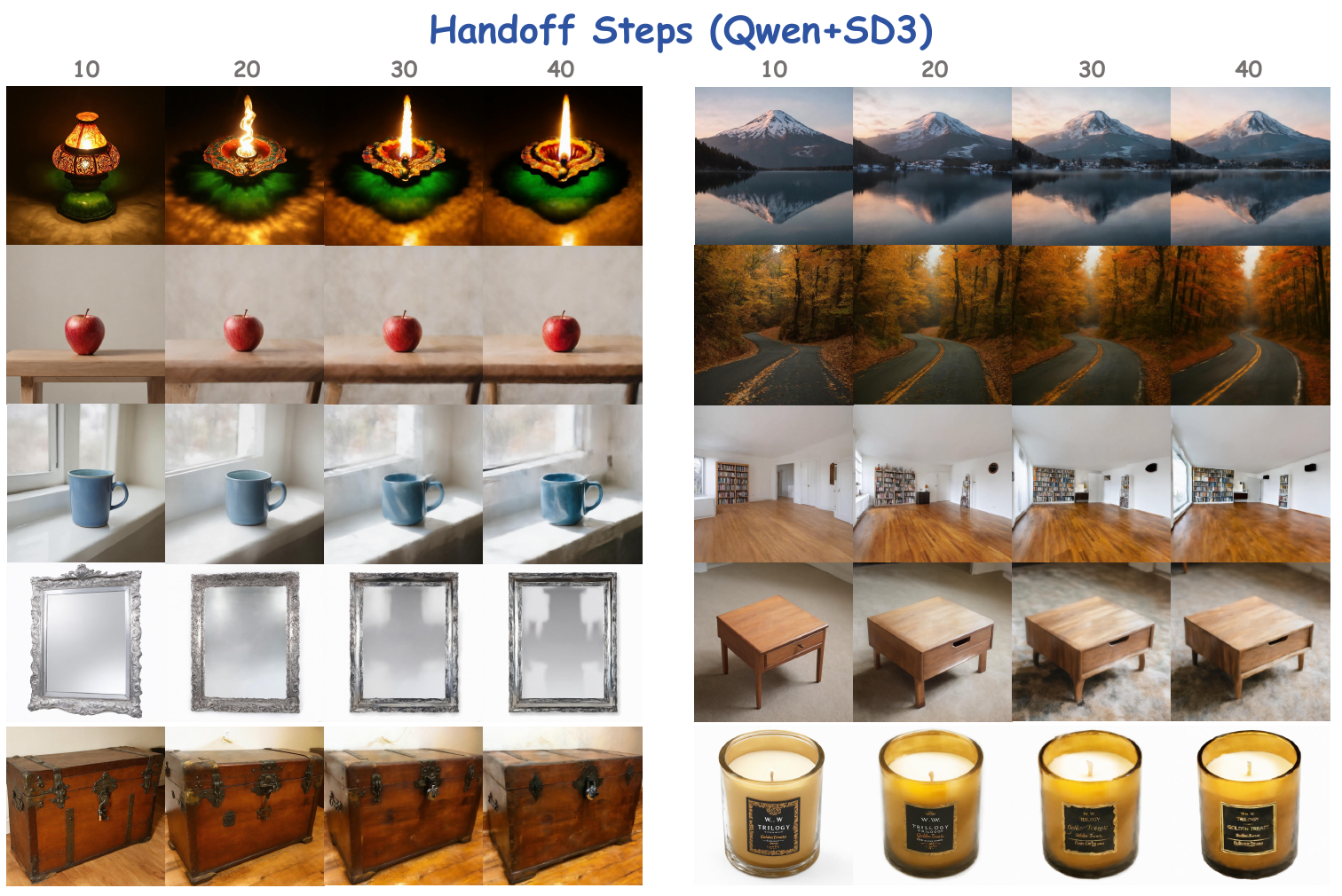}
    \caption{Cross-model state handoffs at different denoising timesteps (Qwen-Image, SD3).}
    \label{fig:app_switch_step_2}
\end{figure}

To evaluate the trajectory-wide alignment of the proposed LAE across the generation process, we conduct cross-model handoffs at arbitrary intermediate timesteps. As depicted in Figs.~\ref{fig:app_switch_step_1} and \ref{fig:app_switch_step_2}, transitions between source and target generators execute reliably at virtually any stage of denoising. This flexibility confirms that our LAE maintains robust state and dynamics alignment throughout the entire trajectory, rather than being restricted to specific timestep windows.

\vspace{-1em}
\subsection{Additional Results across Applications}

\begin{figure}[t]
    \centering
    \includegraphics[width=1\linewidth]{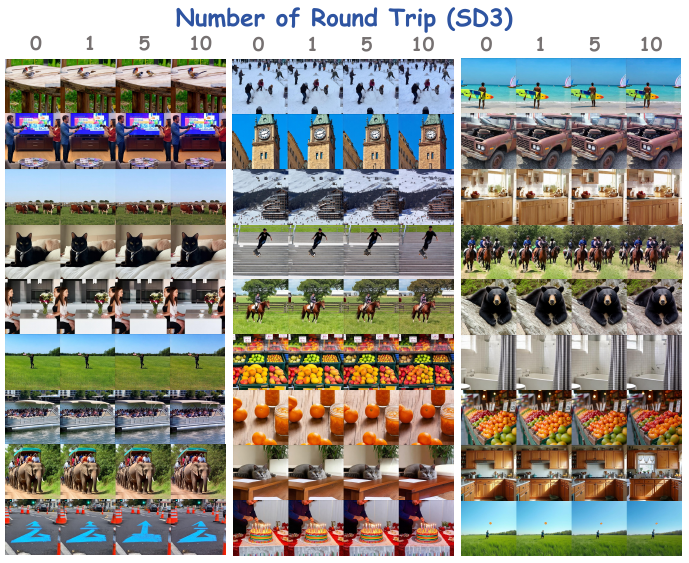}
    \caption{Qualitative comparison of shared latent reconstruction (SD3).}
    \label{fig:app_reconstruct_1}
\end{figure}
\begin{figure}[t]
    \centering
    \includegraphics[width=1\linewidth]{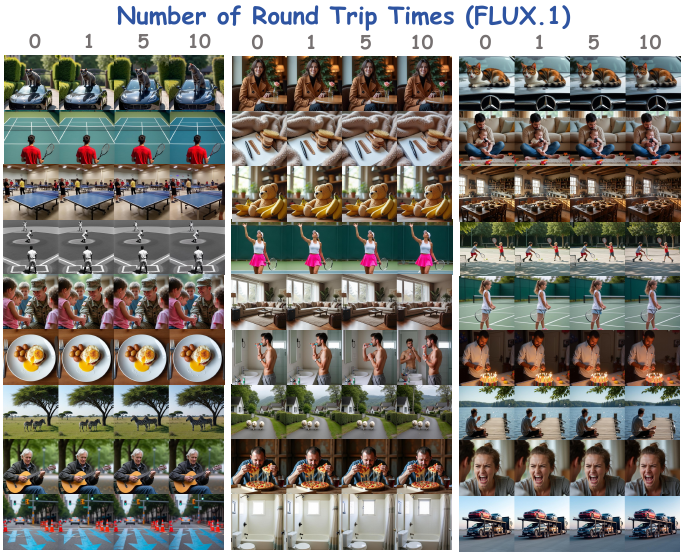}
    \caption{Qualitative comparison of shared latent reconstruction (FLUX.1).}
    \label{fig:app_reconstruct_2}
\end{figure}
\begin{figure}[t]
    \centering
    \includegraphics[width=1\linewidth]{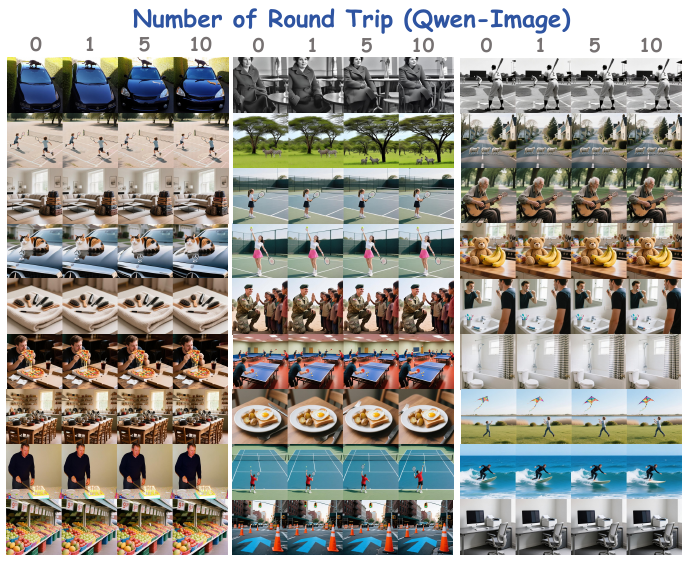}
    \caption{Qualitative comparison of shared latent reconstruction (Qwen-Image).}
    \label{fig:app_reconstruct_3}
\end{figure}
\begin{figure}[t]
    \centering
    \includegraphics[width=1\linewidth]{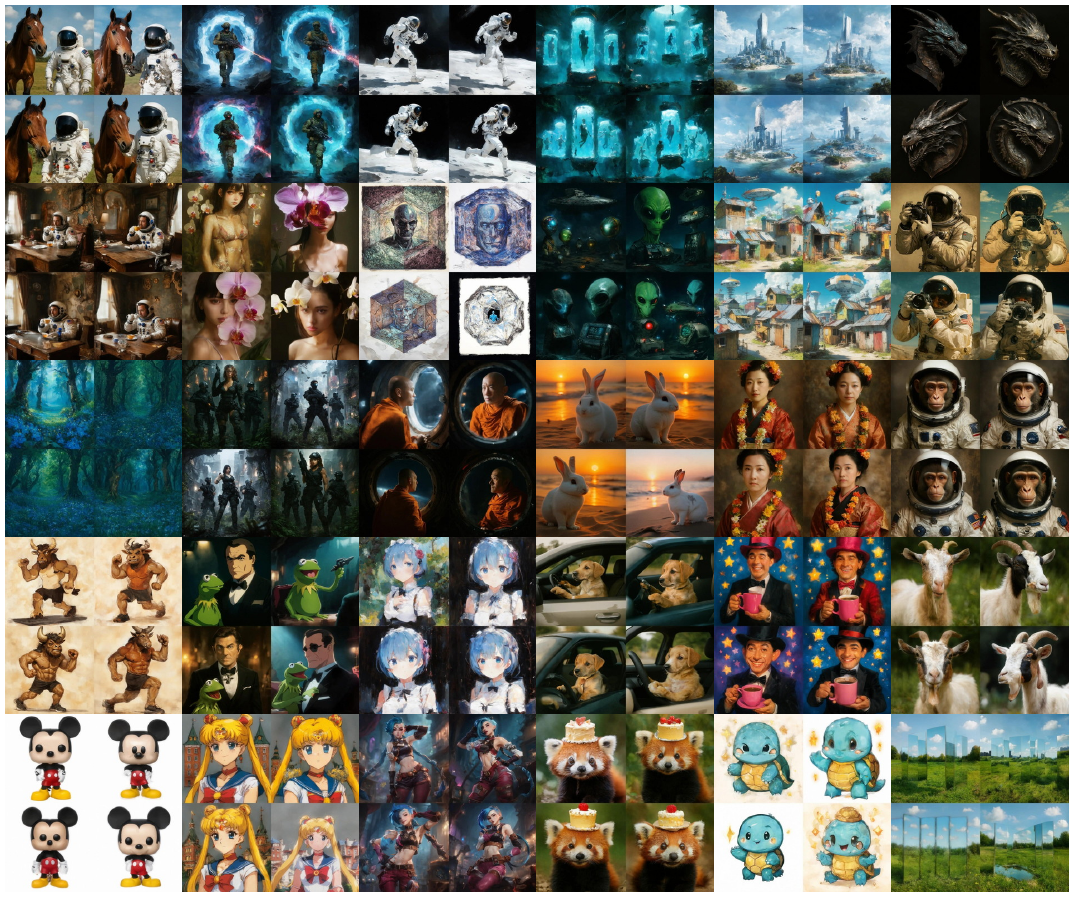}
    \caption{Cross-architecture state handoffs initiated from Qwen-Image to SD3.}
    \label{fig:app_trans_sd3}
\end{figure}
\begin{figure}[t]
    \centering
    \includegraphics[width=1\linewidth]{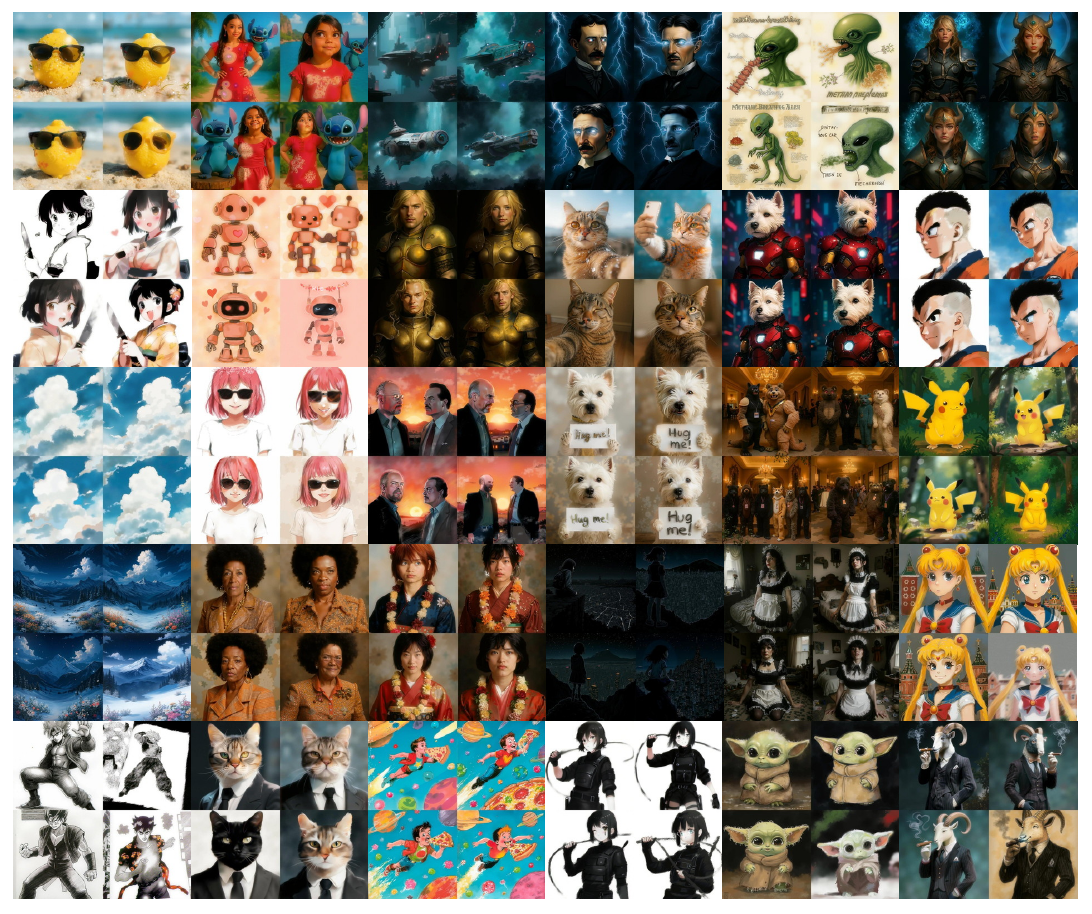}
    \caption{Cross-architecture state handoffs initiated from Qwen-Image to FLUX.1.}
    \label{fig:app_trans_flux}
\end{figure}
\begin{figure}[t]
    \centering
    \includegraphics[width=1\linewidth]{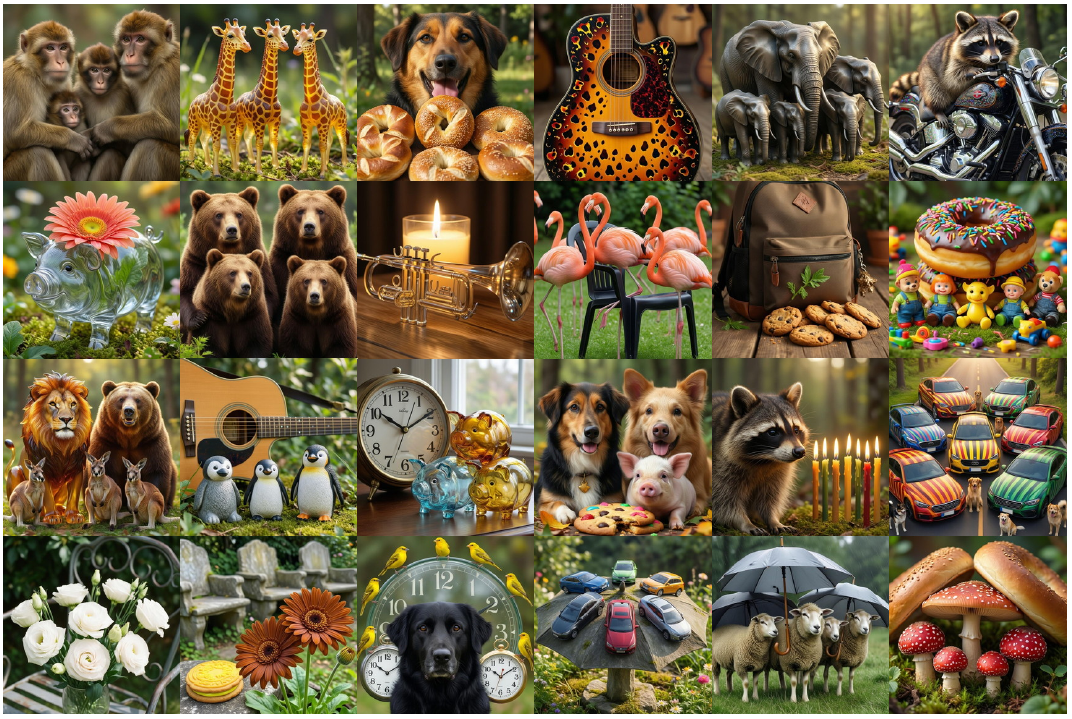}
    \caption{Qualitative results under generation acceleration.}
    \label{fig:app_efficiency}
\end{figure}
\begin{figure}[t]
    \centering
    \includegraphics[width=1\linewidth]{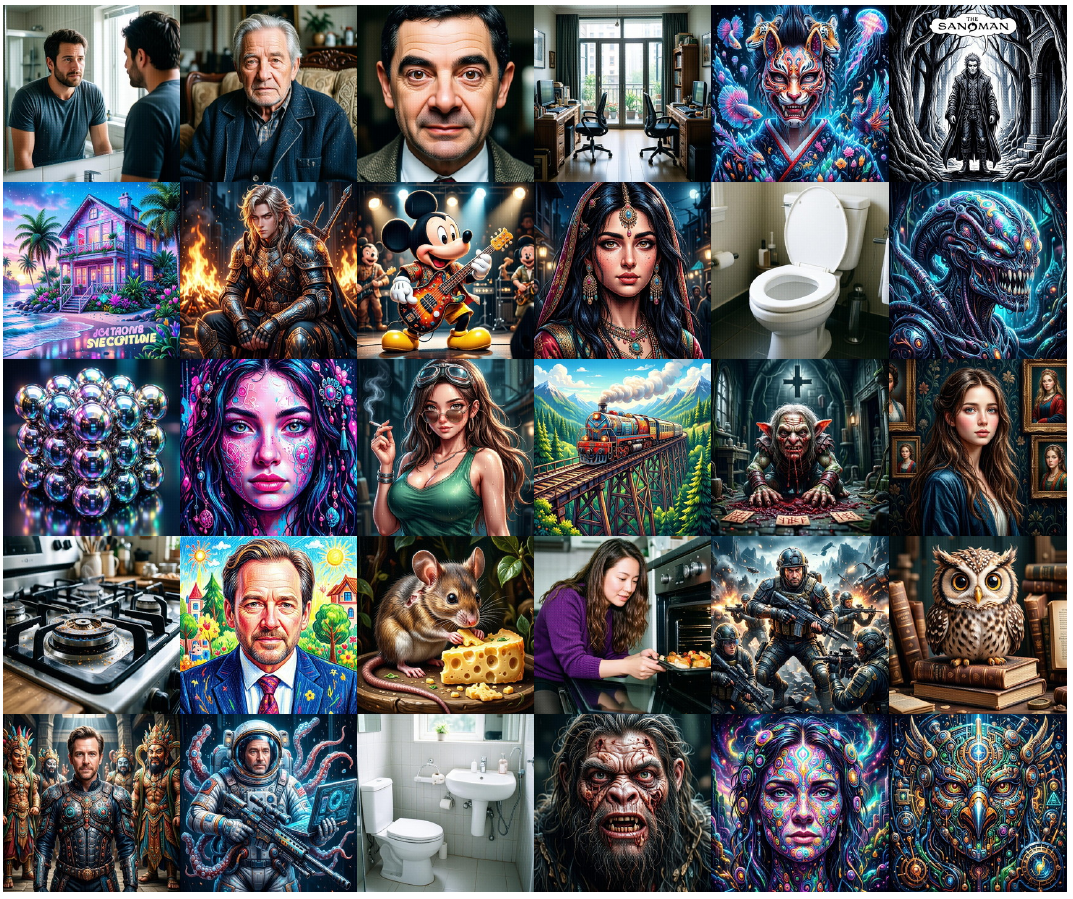}
    \caption{Multi-preference composition cases combining GenEval-v2 and HPSv2.}
    \label{fig:app_multi_pref_geneval_hps}
\end{figure}
\begin{figure}[t]
    \centering
    \includegraphics[width=1\linewidth]{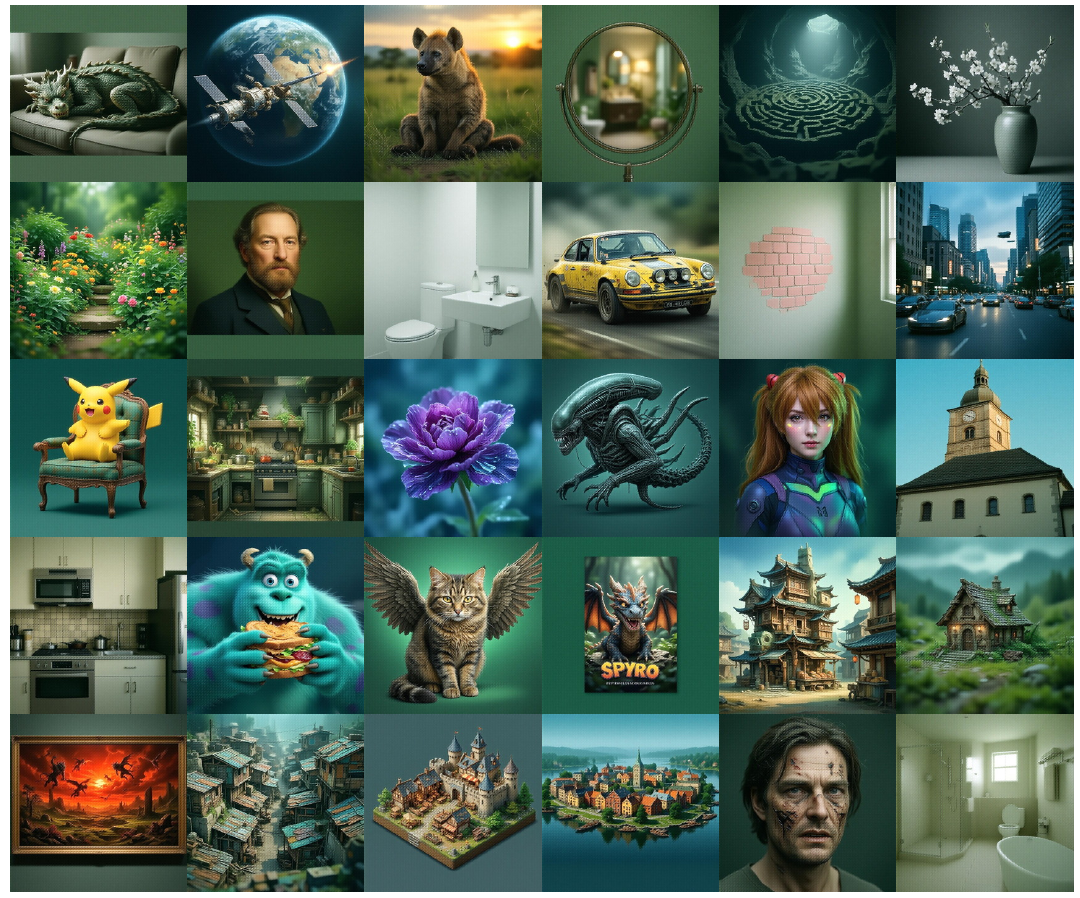}
    \caption{Multi-preference composition cases combining HPSv2 and HPSv3.}
    \label{fig:app_multi_pref_hps_v3}
\end{figure}

To further evaluate the versatility and qualitative fidelity of InstantFusion, this section provides extended visual comparisons across diverse downstream configurations. Specifically, Figs.~\ref{fig:app_reconstruct_1}, \ref{fig:app_reconstruct_2}, and \ref{fig:app_reconstruct_3} depict single-trip and multi-trip latent reconstructions across SD3, FLUX.1, and Qwen-Image, verifying that our shared representation consistently preserves spatial structures across distinct autoencoder designs. Next, Figs.~\ref{fig:app_trans_sd3} and \ref{fig:app_trans_flux} showcase cross-architecture handoffs from Qwen-Image into SD3 and FLUX.1 respectively, highlighting robust state migration between heterogeneous model families. Fig.~\ref{fig:app_efficiency} illustrates the synergistic acceleration. Finally, Figs.~\ref{fig:app_multi_pref_geneval_hps} and \ref{fig:app_multi_pref_hps_v3} demonstrate multi-preference composition results across distinct reward alignments—combining instruction adherence (GenEval-v2) with aesthetic preference (HPSv2), and coupling multi-domain aesthetic metrics (HPSv2 and HPSv3), respectively.


\end{document}